%% file: main.tex
\documentclass[10pt]{IEEEtran}
\include{Packages}
\include{macros}
\usepackage{bbold}

\begin{document}

\title{Tera-SpaceCom: GNN-based Deep Reinforcement Learning for Joint Resource Allocation and Task Offloading in TeraHertz Band Space Networks}

\author{Zhifeng~Hu, Chong~Han,~\IEEEmembership{Senior~Member,~IEEE}, Wolfgang Gerstacker,~\IEEEmembership{Senior~Member,~IEEE}, 
\\and Ian F. Akyildiz,~\IEEEmembership{Life Fellow,~IEEE}
		
		\thanks{
			\par
			Zhifeng Hu is with the Terahertz Wireless Communications (TWC) Laboratory, Shanghai Jiao Tong University, Shanghai 200240, China (email: zhifeng.hu@sjtu.edu.cn).
            \par Chong Han is with Terahertz Wireless Communications (TWC) Laboratory, and with Department of Electronic Engineering and Cooperative Medianet Innovation Center (CMIC), Shanghai Jiao Tong University, Shanghai 200240, China (e-mail: chong.han@sjtu.edu.cn).
			\par Wolfgang Gerstacker is with the Institute for Digital Communications, Friedrich-Alexander-Universität Erlangen-Nürnberg, Erlangen 91058, Germany (email: wolfgang.gerstacker@fau.de).
            \par Ian F. Akyildiz is with Truva Inc., Alpharetta, GA 30022 USA (e-mail: ian@truvainc.com).
		}
}

\maketitle

\input{Abstract}

\input{Introduction}
\input{System}
\input{DRLframework}
\input{GNNDRL} 
\input{Result}

\input{conclusion}

	\bibliographystyle{IEEEtran}
	\bibliography{bibdata}
\end{document}

%% file: Packages.tex
\usepackage{afterpage}
\usepackage{algorithmicx}
\usepackage[linesnumbered,ruled,longend]{algorithm2e}
\SetKwInOut{Input}{Input}
\SetKwInOut{Output}{Output}

\SetKwProg{Fn}{Function}{\string:}{end}

\usepackage{lipsum}
\usepackage{amsfonts} 
\usepackage{amsmath}
\usepackage{amssymb}
\usepackage{array} 

\usepackage{boxedminipage}
\usepackage{caption} 
\usepackage{cite}
\usepackage{color} 
\usepackage{enumerate}
\usepackage{float}
\usepackage{framed}

\usepackage{graphics}
\usepackage{graphicx}
\usepackage{indentfirst}
\usepackage{latexsym}
\usepackage{mathtools}
\usepackage{multirow} 
\usepackage{booktabs}
\usepackage{picinpar}
\usepackage{placeins}
\usepackage{subfigure}
\usepackage{theorem} 
\usepackage[para]{threeparttable}
\usepackage{wrapfig} 
\usepackage{textcomp}
\usepackage{verbatim}
\usepackage{multicol}
\usepackage{stfloats}
\usepackage{soul}

%% file: Abstract.tex
\begin{abstract}
Terahertz (THz) space communications (Tera-SpaceCom) is envisioned as a promising technology to enable various space science and communication applications. 
Mainly, the realm of Tera-SpaceCom consists of THz sensing for space exploration, data centers in space providing cloud services for space exploration tasks, and a low earth orbit (LEO) mega-constellation relaying these tasks to ground stations (GSs) or data centers via THz links. 
Moreover, to reduce the computational burden on data centers as well as resource consumption and latency in the relaying process, the LEO mega-constellation provides satellite edge computing (SEC) services to directly compute space exploration tasks without relaying these tasks to data centers. 
The LEO satellites that receive space exploration tasks offload (i.e., distribute) partial tasks to their neighboring LEO satellites, to further reduce their computational burden. 
However, efficient joint communication resource allocation and computing task offloading for the Tera-SpaceCom SEC network is an NP-hard mixed-integer nonlinear programming problem (MINLP), due to the discrete nature of space exploration tasks and sub-arrays as well as the continuous nature of transmit power. 
To tackle this challenge, a graph neural network (GNN)-deep reinforcement learning (DRL)-based joint resource allocation and task offloading (GRANT) algorithm is proposed with the target of long-term resource efficiency (RE). 
Particularly, GNNs learn relationships among different satellites from their connectivity information. 
Furthermore, multi-agent and multi-task mechanisms cooperatively train task offloading and resource allocation. 
Compared with benchmark solutions, GRANT not only achieves the highest RE with relatively low latency, but realizes the fewest trainable parameters and the shortest running time.
\end{abstract}


%% file: Introduction.tex
\section{Introduction}
\label{sec: intro}

\begin{figure*}[t] 
\centering
        \includegraphics[width=.9\linewidth]{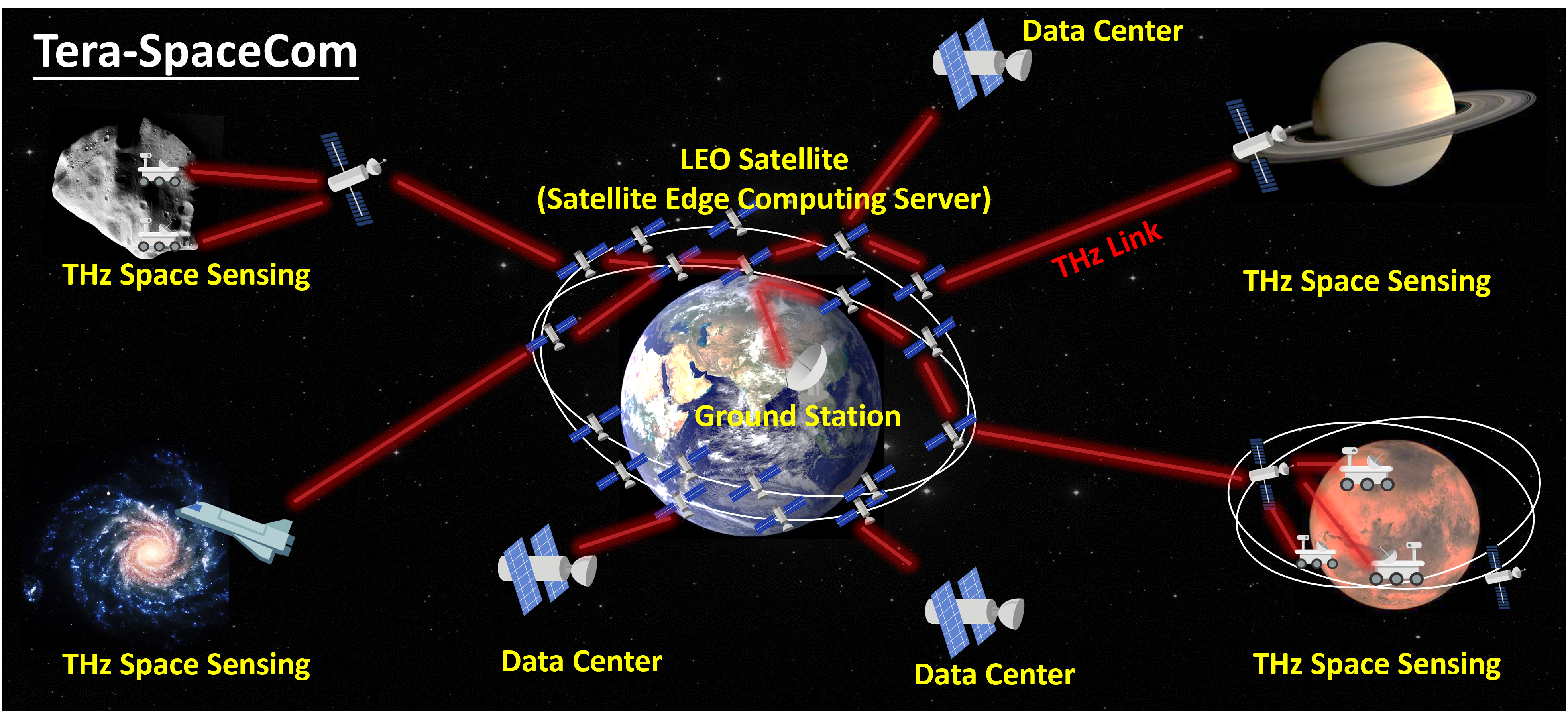} 
        \caption{Tera-SpaceCom.} 
        \label{fig: THz space communication}
\end{figure*}

Terahertz (THz) band (0.1-10 THz) has attracted growing research interests from space science and communication applications, giving rise to an innovative paradigm of THz space communications (Tera-SpaceCom)~\cite{akyildiz2022terahertz}.
First of all, as depicted in Fig.~\ref{fig: THz space communication}, THz sensing is utilized in Tera-SpaceCom for space exploration.
The peak blackbody energy of thermally illuminated gas and dust falls into the THz band for the temperature within 5-100~K~\cite{siegel2010thz}. 
In addition, almost all photons and about half of the luminosity in the universe are emitted within the frequency range of the THz and far infrared bands~\cite{siegel2002terahertz}.
Hence, THz sensing in Tera-SpaceCom is envisioned as a key technology to facilitate various space exploration tasks with respect to planets, star systems, and galaxies~\cite{siegel2007thz}. 
Then, a low earth orbit (LEO) mega-constellation (at altitudes from 500~km to 1500~km) is leveraged in Tera-SpaceCom, which relays the space exploration data to ground stations (GSs).
Unlike medium earth orbit (MEO) and geostationary earth orbit (GEO) satellites with much higher altitudes, the LEO mega-constellation can provide pervasive communication coverage on the Earth~\cite{kak2020designing,kak2019large} with better signal quality and lower propagation delays~\cite{jia2017collaborative}.
In addition, the LEO mega-constellation might relay the data from deep space and near space to the data centers in space~\cite{peng2022integration} for computation and storage.
Compared to data centers on the ground, data centers in space can benefit from the much higher cooling efficiency in space and much larger coverage on the Earth. 

A rapid increase in space exploration tasks~\cite{du2016cooperative} indicates a foreseeable requirement for efficient non-terrestrial transmission and computation solutions with stringent latency requirements.
Although LEO satellites can relay space exploration tasks to data centers 
in space for task computation, the relaying process leads to a substantial computational burden on centralized data centers, as well as large transmission resource consumption and latency.
To reduce the computational overload on data centers~\cite{yu2017survey,zhao2019edge}, LEO satellites can provide satellite edge computing (SEC) services~\cite{denby2020orbital} for space exploration tasks.
In addition, by directly processing the space exploration tasks in LEO SEC satellites without relaying them to data centers via LEO satellites, the latency and transmission resource consumption can be reduced~\cite{jiang2019toward}. 
To further reduce the computational burden on the LEO SEC satellites that receive the space exploration tasks, partial tasks are offloaded from these satellites to their neighboring LEO SEC satellites (i.e., these satellites assign some tasks to neighboring LEO satellites for computation).
To this end, owing to the ultra-broad available bandwidths, the THz band is exploited to support low-latency intersatellite and satellite-ground communications~\cite{akyildiz2022terahertz,hwu2013terahertz}. 

The deployment of THz communications can benefit Tera-SpaceCom from the following five aspects.
\begin{itemize}
\item 
The ultra-broad spectrum resource of THz communications can provide ultra-high date rates and, thus, low transmission delays in space communication networks~\cite{akyildiz2019new}.
\item 
Even though optical communications own ultra-broad bandwidths as well, the very narrow beamwidth poses a quite stringent requirement for beam alignment accuracy, which is challenging in non-terrestrial scenarios with very long propagation distances~\cite{chaccour2022seven}.
In contrast, the wider THz beams ensure a much higher tolerance to beam misalignment~\cite{nie2021channel}.
\item
THz transmissions are not affected by molecular absorption loss in intersatellite communications, further improving the data rates and reducing delays. 
\item
Due to the sub-millimeter wavelength, THz transceivers are capable of utilizing an ultra-large-scale antenna array, which consists of a large number of sub-millimeter-long antennas.
Therefore, multiple-input and multiple-output (MIMO) and hybrid beamforming technologies can be applied to further improve the spectral efficiency~\cite{han2021hybrid}.
\item 
The combination of THz communications and aforementioned THz sensing follows the trend of integrated communication and sensing (ISAC) in space~\cite{aliaga2022joint}.
\end{itemize}

Despite the promising potential of Tera-SpaceCom, resource-efficient computing task offloading and communication resource allocation in Tera-SpaceCom SEC networks are still challenging.
This challenge is the focus of this paper, while THz sensing and deep space communications are not the subject of this work. 
On the one hand, efficient sub-array management for hybrid beamforming, as well as expenses brought by transmit power consumption, are crucial concerns in practical THz systems~\cite{zhai2019antenna,hu2023deep}. 
Therefore, sub-array and transmit power resource usage need to be minimized, for the purpose of resource-efficient transmission in Tera-SpaceCom.
On the other hand, to reduce the resource usage in the end-to-end computation and transmission process for Tera-SpaceCom SEC services, task offloading and resource allocation can be jointly considered.
More specifically, the task offloading determines the amount of tasks offloaded to different LEO satellites, while resource allocation decides the amount of transmit power and sub-arrays in each LEO satellite to support the transmission of computing tasks and outcomes.
However, due to the continuous nature of transmit power and the discrete nature of sub-arrays and tasks, the joint task offloading and power and sub-array allocation problem is a mixed-integer nonlinear programming problem (MINLP)~\cite{li2023graph}.
Hence, this problem is non-convex and NP-hard, especially in the LEO mega-constellation with a large number of involved satellites and stochastic computing demands. 

Recent research efforts have been devoted to resource efficiency (RE, i.e., the computation and transmission rate per resource usage) the maximization by considering both resource allocation and task offloading in SEC networks, which is one of the key components of Tera-SpaceCom. 
Non-learning solutions maximize RE based on iterative optimization methods (i.e., iterative Lagrange multiplier method~\cite{ding2021joint, song2021energy, leyva2023satellite}  and iterative interior-point method~\cite{ding2023dynamic}).
Nevertheless, for any time instant, these algorithms need multiple iterations with prohibitively high complexity, which is not applicable in highly dynamic environments of mega-constellation systems~\cite{wang2022joint}.

Moreover, owing to the recent rapid development of machine learning, reinforcement learning (RL) and deep reinforcement learning (DRL) based solutions are leveraged in~\cite{shi2024joint,chai2023joint,huang2024joint} for long-term RE maximization.
Different from non-learning methods, RL and DRL solutions can provide task offloading and resource allocation without time-consuming iterative computation for every time instant.
The RL solution~\cite{shi2024joint} is a tabular method that collects the accumulated values of RE and latency in a long-term period for each pair of wireless environment state and offloading action.
However, such a tabular method is impractical in complicated mega-constellation networks with very large discrete or even infinite continuous state and action spaces.
On the contrary, by adopting the neural networks with powerful learning abilities into RL, DRL methods~\cite{chai2023joint,huang2024joint} can estimate optimal actions for unseen states, ensuring their applicability in complex environments.
However, these methods focus on computing resources, while communication resource allocation is not taken into account.

In addition, the learning-based methods, along with the non-learning algorithms, all require that stochastic amounts of tasks for all time instants are known.
This prerequisite is unrealistic in mega-constellation, since after the transmission of environment state, as well as training or task offloading and resource allocation calculation, the amount of tasks used for training or calculation might be obsolete.
Furthermore, the sub-array management is ignored by existing works, which is vital for THz networks with large-scale antenna arrays. 

For the purpose of addressing the aforementioned issues, we propose a graph neural network (GNN)-DRL based joint resource allocation and task offloading (GRANT) algorithm in SEC networks of Tera-SpaceCom.
More concretely, we utilize GNNs as the neural network architecture of DRL to offload the tasks, as well as allocate transmit power and sub-arrays, for the target of long-term RE maximization.
The main contributions in this work are summarized as follows.
\begin{itemize}
    \item In the Tera-SpaceCom SEC network on the basis of LEO mega-constellation, we formulate a joint communication resource allocation and computing task offloading problem.
    Specifically, the target of the formulated problem is to maximize the long-term RE in the entire Tera-SpaceCom SEC network from the perspective of transmit power and sub-array utilization. 
    Physical constraints are further taken into consideration pertaining to the maximal power and number of sub-arrays in each LEO satellite, as well as the low-latency requirement.
    \item We transform the joint resource allocation and task offloading problem into an equivalent DRL problem, with the customized design of key DRL components, namely, state, action, and reward.
    The proposed definition of state enables DRL to work without knowing the exact computing demands, making DRL resilient to the stochastic amount of computing tasks.
    In addition, the tailored action and reward alleviate the physical constraint violation.
    \item We design a GRANT algorithm to solve the DRL problem. 
    In GRANT, GNNs are leveraged to capture the relationship among neighboring satellites via the connectivity structure of the LEO mega-constellation. 
    Moreover, unnecessary input states and actions are pruned away, which cannot only make GRANT focus on useful input, but reduce the computational burden of GRANT. 
    In addition, the multi-agent and multi-task mechanisms allow collaborative training for task offloading and resource allocation among all LEO satellites.
    One step further, safe initialization and safe exploration mechanisms are adopted to mitigate large latency during training.
    \item We assess the performance of our proposed GRANT algorithm.
    Remarkably, GRANT realizes the lowest resource usage (i.e., the highest RE) in a mega-constellation with 105~ms latency, while the benchmark solutions either have more than 33\% larger latency, or achieve 20\% less latency at the expense of 83\% more resource usage.
    Moreover, THz communications realize only $\frac{1}{43}$ and $\frac{1}{197}$ of the latency of low-frequency Ka and Ku bands, respectively.
    In addition, compared to other algorithms, GRANT has the lowest number of trainable parameters and running time, suggesting its high memory efficiency and computational efficiency.
\end{itemize}

The remainder of the paper is structured as follows.
We introduce the SEC model in Tera-SpaceCom, as well as formulate the long-term joint task offloading and resource allocation problem in Sec.~\ref{sec: system}.
We depict a DRL framework for the formulated problem in Sec.~\ref{sec: DRL description}.
The GRANT algorithm is elaborated in Sec.~\ref{sec: GNN-DRL}.
Simulation results are presented and discussed in Sec.~\ref{sec: result}.
Finally, this paper is concluded in Sec.~\ref{conclusion}.

%% file: System.tex
\section{System Model and Problem Formulation}
\label{sec: system}

In this section, we take into account an SEC network in Tera-SpaceCom, which consists of a LEO satellite mega-constellation and a GS.
Based on this SEC model, we formulate a joint computing task offloading and communication resource allocation problem for long-term RE maximization.

\subsection{Tera-SpaceCom SEC Network Model}
\label{sec: satellite model}

We consider an SEC network in Tera-SpaceCom, based on a LEO satellite mega-constellation, which is composed of multiple LEO satellites located in multiple orbits.
Each LEO satellite in the mega-constellation is equipped with the same type of processing unit, which serves as an SEC server.
The GS requests data for multiple space exploration tasks, which require computing services (e.g., data compression, image processing, etc.).
Multiple LEO satellites in the Tera-SpaceCom mega-constellation first receive these computing tasks.
To reduce the computing burden of centralized data centers, as well as reduce the latency of relaying tasks from LEO satellites to data centers, some space exploration tasks are directly processed by LEO satellites. 
In this paper, we focus on these tasks computed by Tera-SpaceCom LEO SEC servers.
Then, the LEO satellites that receive the unprocessed tasks offload some tasks to other LEO satellites in the Tera-SpaceCom mega-constellation via THz communications.
After the SEC servers finish the computation, the computing results are eventually transmitted to the GS by leveraging THz communications.

In this paper, we adopt the commonly used Walker constellation structure~\cite{wood2003satellite,jia2017collaborative,su2019broadband}, which has been successfully applied to commercial constellations to provide global coverage, e.g., Iridium, Galileo, etc. 
In particular, the Walker constellation consists of multiple circular orbits (or planes) with the same inclination and altitude.
Each orbit accommodates the same number of satellites, while these satellites are equally spaced in this orbit.
Moreover, the orbital planes are equally spaced around the equator.

In this mega-constellation, the LEO satellites that receive unprocessed tasks can offload these tasks to neighboring LEO satellites~\cite{zhang2022aerial,cassara2022orbital} through THz intersatellite links (ISLs), with the following benefits.
On one hand, task offloading can reduce the computational burden and time compared to centralized computation.
On the other hand, since the data volume of tasks is remarkably reduced after computation~\cite{chen2018joint,chen2015efficient}, the subsequent data transmission burden can be alleviated by processing the tasks via neighboring satellites.
In addition, each LEO satellite can establish 4 ISLs with neighboring satellites simultaneously, which is commonly adopted in both academia~\cite{liu2020drl} and industry (e.g., Iridium and Starlink)~\cite{liu2020drl,chaudhry2021laser}.
Specifically, the satellite can support two intra-orbit ISLs and two inter-orbit ISLs.
The intra-orbit ISLs link each satellite with its two neighboring satellites in the same orbit, while the inter-orbit ISLs connect the satellite with its neighboring satellites in the two adjacent orbits that have the closest orbit phasing (i.e., true anomaly).
Benefiting from the 4 ISLs with adjacent neighbors, the spreading loss brought by the propagation distance is lower, compared to the communications with farther satellites.
Furthermore, the LEO satellites communicate with each other in a mesh manner, enabling flexible routing for data transmission. 

The unprocessed task data offloaded to an SEC server $i$ is computed by the processing unit inside this SEC server.
The computational time $T^{cp}$ for time slot $\tau$ is given by
\begin{equation}
    \label{eq: computation delay}
    T^{cp}_\tau(i)=\dfrac{L_\tau(i)z}{Q},
\end{equation}
where $L_\tau(i)$ is the amount of computing task data in byte offloading to the SEC server $i$, $z$ represents the number of CPU cycles required for each byte of data, and $Q$ refers to the computation ability of the server in CPU cycles per second.
After the computation is completed, the size of the outcome $L'_\tau(i)$ is
\begin{equation}
    \label{eq: outcome size}
    L'_\tau(i)=\beta L_\tau(i),
\end{equation}
where $\beta$ is a factor less than 1 in this paper, since the size of the outcome is less than the size of the unprocessed data~\cite{chen2018joint,chen2015efficient}.

After computation, the processed tasks are transmitted to the GS.
The GS first accesses the closest LEO satellite among the satellites it can communicate with (i.e., satellites above its minimum elevation angle).
Then, by taking into account the stability of the wireless connection, GS maintains communications with this satellite until this satellite is not accessible anymore, and then GS establishes a link with the new nearest satellite at this moment.
Computing results of other SEC servers are aggregated in the satellite connected to GS.
For ease of the objective of RE maximization, the satellite communication topology follows the resource-efficient routing designs in~\cite{hu2023deep}, which reduces the resource consumption by jointly considering the hop count and propagation loss of THz waves.
Since SEC servers can conduct both the task data storage and computation~\cite{cao2020overview}, each LEO satellite deploys a first-in-first-out (FIFO) buffer for each ISL link.
In particular, if a satellite relays the computing outcome from multiple satellites, it sends these results according to the order of arrivals.

\subsection{Tera-SpaceCom Transmission Model}
\label{sec: transmission model}

The data transmission for the satellite edge computing task is composed of two phases, namely, the task offloading phase and the outcome transmission phase.
In the task offloading phase, the unprocessed tasks are offloaded to multiple SEC servers from the SEC servers that receive tasks from satellites outside the LEO mega-constellation
Then, in the outcome transmission, the computing outcomes are transmitted to the GS after satellites accomplish computation.
To support the signal transmissions in space, $K$ sub-bands in the THz band are leveraged.
In particular, partial sub-bands are utilized to offload the computing tasks, while the rest support the outcome transmission.

To enlarge the signal strength as well as mitigate interference caused by other links, hybrid beamforming is utilized for both ISLs and satellite-ground links~\cite{cao2007multihop}.
In particular, we adopt the dynamic hybrid beamforming architecture~\cite{yan2020dynamic} with single-stream transmission for each beam.
Such architecture is specifically designed for THz communications and enables the flexible allocation of multiple sub-arrays in transceivers.
Consequently, for the azimuth and elevation angles of transmitted or received signal $\phi$ and $\theta$, the array steering vector is given by
\begin{equation}
\begin{aligned}
    \label{eq:steeringVector}
    \mathbf{a}(\phi, \theta)=&\dfrac{1}{\sqrt{M_x M_y}}\left[1,\dots,e^{j\frac{2\pi d_0}{\lambda}(m_x\sin(\theta)\cos(\phi)+m_y\cos(\theta))},\right.\\
    &\quad \dots,\left.e^{j\frac{2\pi d_0}{\lambda}((M_x-1)\sin(\theta)\cos(\phi)+(M_y-1)\cos(\theta))}\right]^T,
\end{aligned}
\end{equation}
where $d_0$ represents the spacing of neighboring antennas, and $\lambda$ denotes the signal wavelength.
Each sub-array is a planar array with $M_x\times M_y$ antennas.
Additionally, $0\leq m_x \leq M_x-1$, and $0\leq m_y \leq M_y-1$. 

For the data transmission from a satellite $i$ to another satellite (or GS) $j$, the MIMO channel response for the $k^{\text{th}}$ sub-band in time slot $\tau$ is computed as~\cite{peng2021hybrid}
\begin{equation}
\begin{aligned}
    \label{eq:MIMOresponse}
    &\mathbf{H}_\tau(i,j,k)=\\
    &\sqrt{(S_{Tx,\tau}(i,j)M_x M_y)(S_{Rx,\tau}(j,i)M_x M_y)}G_{Tx}G_{Rx}e^{\tilde{j}2\pi f_d \tau \Delta \tau}
    \\&\cdot \mathbf{a}_{Rx}(\phi_{Rx,\tau},\theta_{Rx,\tau})\mathbf{a}_{Tx}^*(\phi_{Tx,\tau},\theta_{Tx,\tau})\alpha_\tau(i,j,k),
\end{aligned}
\end{equation}
where $S_{Tx,\tau}(i,j)$ and $S_{Rx,\tau}(j,i)$ represent the numbers of sub-arrays for satellite $i$ (i.e., transmitter) and satellite (or GS) $j$ (i.e., receiver), respectively.
$G_{Tx}$ and $G_{Rx}$ are the antenna gains for the transmitter and receiver.
$\Delta \tau$ symbolizes the duration of a time slot.
$f_d$ is the Doppler frequency~\cite{al2024doppler}.
$\tilde{j}$ stands for the imaginary unit.
According to $\phi_{Tx,\tau}$/$\theta_{Tx,\tau}$ and $\phi_{Rx,\tau}$/$\theta_{Rx,\tau}$ referring to the azimuth/elevation angles of departure and arrival (AoD/AoA), $\mathbf{a}_{Tx}$ and $\mathbf{a}_{Rx}$ stand for steering vectors of the transmitter and receiver, as given in~\eqref{eq:steeringVector}.
$(\cdot)^*$ is the conjugate transpose operator.
Furthermore, $\alpha_\tau(i,j,k)$ denotes the path gain of THz LoS channel, which is given by
\begin{equation}
\label{eq:pathloss}
  |\alpha_\tau(i,j,k)|^2=
  \begin{aligned}
  &\left(\dfrac{c}{4\pi f d_\tau(i,j)} \right)^2e^{\int_{d_{i,\tau}}^{d_{j,\tau}}{\left(-g_{abs}(f_k,d')\right)}\mathrm{d} d'},
  \end{aligned}
\end{equation}
where $c$ represents the speed of light, $d_\tau(i,j)$ measures the distance between satellite $i$ and satellite (or GS) $j$, $d_{i,\tau}$ and $d_{j,\tau}$ denote the 3D location vectors of $i$ and $j$, $f_k$ refers to the carrier frequency of the $k^{\text{th}}$ sub-band, and $g_{abs}(f_k,d')$ represents the medium absorption coefficient of THz signals for satellite communications~\cite{yang2024universal}.
The channel response corresponding to \eqref{eq:MIMOresponse} including beamforming is
\begin{equation}
\begin{aligned}
   h_\tau(i,j,k)=&
    \mathbf{C_D^*}_{,\tau}(j,i,k)\mathbf{C_A^*}_{,\tau}(j,i,k)\mathbf{H}_\tau(i,j,k)\\
    &\cdot \mathbf{W_{A}}_{,\tau}(i,j,k)\mathbf{W_{D}}_{,\tau}(i,j,k),
\end{aligned}
\label{eq:SISOresponse} 
\end{equation}
where $\mathbf{W_{A}}(i,j,k)$ and $\mathbf{W_{D}}(i,j,k)$ represent the analog and digital precoding matrices, while $\mathbf{C_A^*}(j,i,k)$ and $\mathbf{C_D^*}(j,i,k)$ stand for the analog and digital combining matrices. 

As a result, the signal-to-interference-plus-noise ratio (SINR) of the $k^\text{th}$ sub-band is computed as
\begin{equation}
\gamma_\tau(i,j,k)=\frac{P_\tau(i,j,k)|{h}_\tau(i,j,k)|^2}
{I_{s,\tau}(i,j,k)+||\mathbf{C_D^*}_{,\tau}(j,i,k)\mathbf{C_A^*}_{,\tau}(j,i,k)||^2\sigma^2},
\label{eq:gamma}
\end{equation}
where $P_\tau(i,j,k)$ denotes the transmit power assigned to the $k^\text{th}$ sub-band from node $i$, and $I_{s,\tau}(i,j,k)$ indicates the interference after the self-interference cancellation and beamforming, which can be modeled as a Gaussian variable~\cite{lei2020deep,zhang2019joint}.
$\sigma^2$ is the noise variance.

Therefore, the channel capacity for the transmission from satellite $i$ to $j$ is given by
\begin{equation}
	\label{eq:rate}
R_\tau(i,j)=\sum\limits_{k=1}^{K}\psi_\tau(i,j,k)B\log_2(1+\gamma_\tau(i,j,k)),
\end{equation}
where $\psi_\tau(i,j,k)$ is 1 or 0, indicating whether the $k^\text{th}$ sub-band is utilized for the transmission from node $i$ to $j$ or not.
$B$ denotes the bandwidth for each sub-band.
Benefiting from the beamforming solution in~\cite{yan2020dynamic}, the optimal precoding and combining matrices ($\mathbf{W_{A}}$, $\mathbf{W_{D}}$, $\mathbf{C_A^*}$, and $\mathbf{C_D^*}$) are reachable.

The delay for communication results from the summation of two parts, namely, the transmission delay and the propagation delay~\cite{liu2022energy}.
More concretely, the transmission delay from satellite $i$ to satellite or GS $j$ for a transmission path $\mathcal{L}$ is determined by the transmission rate $R_\tau(i,j)$ in~\eqref{eq:rate}, the amount of data required to be sent $L^\mathcal{L}_\tau(i,j)$, as well as the task data amount $\imath^\mathcal{L}_\tau(i,j)$ in the buffer that should be sent before $L^\mathcal{L}_\tau(i,j)$, which is expressed as 
\begin{equation}
    \label{eq: transmission delay}
    T^{tr,\mathcal{L}}_\tau(i,j)=\dfrac{L^\mathcal{L}_\tau(i,j)+\imath^\mathcal{L}_\tau(i,j)}{R_\tau(i,j)}.
\end{equation}
In addition, the propagation delay is determined by the distance between the transceivers $d_\tau(i,j)$ and the speed of light $c$, which is computed as
\begin{equation}
    \label{eq: propagation delay}
    T^{pr}_\tau(i,j)=\dfrac{d_\tau(i,j)}{c}.
\end{equation}

As a result, the overall delay along the transmission path $\mathcal{L}$ contains the computing delay, as well as the transmission delay and propagation delay, which is given by~\cite{liu2023online}
\begin{equation}
    \label{eq: one path delay}
    T^\mathcal{L}_\tau (i_0)=T^{cp}_\tau(i_1)+\sum\limits_{(i,j)\in \mathcal{L}}(T^{tr,\mathcal{L}}_\tau(i,j)+T^{pr}_\tau(i,j)),
\end{equation}
where $i_0\in\mathbf{N}^\circ$, and $\mathbf{N}^\circ$ denotes the set of LEO SEC servers that receive the unprocessed computing tasks before task offloading.
$\mathcal{L}\in\mathbb{L}_{i_0}$, where $\mathbb{L}_{i_0}$ is the set of paths starting from the LEO satellite $i_0$.
The LEO satellite $i_1$ is the SEC server that processes the tasks, which is on the path $\mathcal{L}$.
The overall delay for the unprocessed tasks received by the LEO satellite $i_0$ is determined by the largest delay of the transmission paths $\mathbb{L}_{i_0}$ starting from $i_0$, which is computed as
\begin{equation}
    \label{eq: one dc delay}
    T^{o}_\tau(i_0)=\max\limits_{\mathcal{L}\in\mathbb{L}_{i_0}}T^\mathcal{L}_\tau(i_0).
\end{equation}

\subsection{Joint Resource Allocation and Task Offloading Problem Formulation}
\label{sec: problem formulation}

The objective of the SEC network in Tera-SpaceCom is to optimize the policy $\pi_p$ of joint computing task offloading (denoted by $\mathbf{L}$) and resource allocation (i.e., the power and sub-array allocation represented by $\mathbf{P}$ and $\mathbf{S}$, respectively), to realize the long-term RE maximization in the Tera-SpaceCom SEC network.
Under the given computing demands of space exploration tasks, the RE maximization is equivalent to the resource usage minimization.
As aforementioned, since the LEO satellites are equipped with the same type of processing unit, the energy consumption caused by computation is fixed for a given amount of tasks~\cite{zhu2023collaborative}. 
Hence, the RE maximization in this paper focuses on communication resource usage. 
More specifically, the usage of transmit power and sub-arrays are taken into consideration, which are commonly considered in RE maximization in THz wireless communication systems~\cite{zhai2019antenna,hu2023deep}.

In particular, the power consumption ratio for a LEO satellite $i\in\{1,2,\dots,N\}$ is defined as \begin{equation}
    \label{eq:power occupation}
    U_{P,\tau}(i)=\sum\limits_{j\in\{1,2,\dots,N+1\}\backslash\{i\}}\sum\limits_{k\in\{1,2,\dots,K\}} \dfrac{\psi_\tau(i,j,k)P_\tau(i,j,k)}{P_\text{max}},
\end{equation}
where $1,2,\dots,N$ denotes the indices of LEO satellites that are involved in the computation and transmission, while $N+1$ is the index of GS.
$P_\text{max}$ refers to the maximal transmit power for a LEO satellite.

In addition, to ensure each satellite can receive signals from different neighboring satellites simultaneously with alleviated interference, each satellite is equipped with one unique receiving sub-array for each ISL.
By contrast, the number of transmitting sub-arrays depends on resource allocation decisions.
As a result, the transmitting sub-array occupation ratio is expressed as 
\begin{equation}
    \label{eq:array occupation}
    U_{S,\tau}(i)=\sum\limits_{j\in\{1,2,\dots,N+1\}\backslash\{i\}}\dfrac{S_{Tx,\tau}(i,j)}{S_{\max}},
\end{equation}
where $S_\text{max}$ stands for the maximal number of transmitting sub-arrays in a LEO satellite. 

The overall resource usage is the weighted summation of utilized power and sub-array ratios.
To emphasize RE in terms of both power and sub-array resources, equal weights are assigned to power and sub-array ratios, which enables a balanced trade-off between the usage of different resources~\cite{hu2023deep,tang2014resource}.
Therefore, the overall resource usage $U_\tau(i)$ for a LEO satellite $i$ can be expressed as the mean of the power and sub-array occupations, given by
\begin{equation}
    \label{eq: total consumption}
    U_\tau(i)=\dfrac{U_{P,\tau}(i)+U_{S,\tau}(i)}{2}.
\end{equation}

To formulate the RE maximization problem, the objective is to minimize the expected resource consumption (i.e., the resource consumption averaged over all the satellites that offload computing tasks, as well as all the satellites that transmit computing outcomes) of the Tera-SpaceCom SEC network for a long-term period (i.e., spanning over multiple time slots $\tau$ starting from any time instant $t$), as
\begin{subequations}
        \label{eq:total objective}
        \begin{align}
        \label{eq: object}
        &\mathop{\arg\min}\limits_{\pi_p\left({\mathbf{L}_\tau, \mathbf{S}_\tau,\mathbf{P}_\tau}\right)}
        \sum\limits_{\tau=t}^\infty \kappa^{\tau-t} \mathbb{E}_{\pi_p}\left[U_{\tau}\right],\\
        \label{eq: task constraint}
        \textrm{s.t.}\quad 
        & L_\tau(i)+\sum\limits_{j\in \mathcal{N}^i}L_\tau(i,j)=L^\circ_\tau (i), \forall i\in \mathbf{N}^\circ, \tau, \\
        \label{eq: power constraint}
        &\sum\limits_{j\neq i}\sum\limits_{k} \psi_\tau(i,j,k)P_\tau(i,j,k)\leq P_\text{max},\forall i,\tau, \\
        \label{eq: subarray constraint}
        &\sum\limits_{j\neq i}S_{Tx,\tau}(i,j)\leq S_\text{max},\forall i,\tau, \\
        \label{eq: delay constraint}
        &T^{o}_\tau\leq T_{\text max}, \forall \tau.
        \end{align}
\end{subequations}
In~\eqref{eq: object}, $\kappa\in[0,1]$ is the discount factor for future resource usage.
Specifically, a larger $\kappa$ represents a larger weighted contribution of future resource occupation.
On the contrary, a smaller $\kappa$ highlights the current resource occupation. 
$U_{\tau}$ denotes the overall resource usage averaged over involved LEO satellites in both the task offloading phase and the outcome transmission phase. 
In~\eqref{eq: task constraint}, for a LEO satellite that offloads tasks, the summation of offloaded data amounts equips to the unprocessed data amount arriving at this LEO satellite.
$\mathcal{N}^i$ refers to the set of satellites that have ISLs with LEO satellite $i$.
In the task offloading phase, the LEO satellite $i$ and its neighbors in $\mathcal{N}^i$ are the satellites that process the computing tasks offloaded by the satellite $i\in\mathbf{N}^\circ$ (as aforementioned, $\mathbf{N}^\circ$ denotes the set of SEC servers that receive unprocessed computing tasks before task offloading).
$L_\tau^\circ(i)$ is the amount of task data that arrived at LEO satellite $i$ before task offloading.
As shown in~\eqref{eq: power constraint} and~\eqref{eq: subarray constraint}, the transmit power and the total used sub-arrays for any involved LEO satellites $i$ that sends or relays computing tasks or outcome cannot exceed the maximal values that this satellite can provide.
Furthermore, the latency $T^{o}_\tau$ averaged over all the LEO satellites in $\mathbf{N}^\circ$ cannot exceed the limitation, formulated as~\eqref{eq: delay constraint}.

%% file: DRLframework.tex
\section{Deep Reinforcement Learning Framework}
\label{sec: DRL description}

In this section, we establish a DRL framework to solve the long-term joint task offloading and resource allocation problem in the Tera-SpaceCom SEC network.
On the basis of the observed state of the wireless environment (i.e., the Tera-SpaceCom SEC network), two DRL learning agents generate the task offloading and resource allocation actions in the task offloading phase and outcome transmission phase, respectively.
On the one hand, in the task offloading phase, the first DRL agent determines the amount of offloaded computing tasks, as well as the allocation of communication resources to transmit these unprocessed tasks.
On the other hand, the second DRL agent allocates the communication resources for the links that send the processed data to the GS in the outcome transmission phase.
Particularly, the DRL agents are located in the GS, which collects the wireless environment state of the entire Tera-SpaceCom SEC system.

Particularly, the overall DRL optimizes the task offloading and resource allocation actions as follows:
First, two DRL agents obtain the observed state $s_\tau$ from the LEO mega-constellation (including the estimated SINR, the mean data amount to be processed and transmitted, the index that indicates the locations of the satellite, which satellites offload computing tasks, and which satellite is connected to GS).
By processing the observed state via the current trained policy $\pi_p$, the agents determine the task offloading and communication resource allocation actions $\Xi_\tau$.
Accordingly, the observed state is updated from $s_\tau$ to $s_{\tau+1}$.
Furthermore, the reward $r_\tau$ can be obtained in Tera-SpaceCom, which is related to the resource consumption in the objective~\eqref{eq:total objective}.
By learning from the experience of actions and rewards, the agents update the policy to select actions with the highest accumulated rewards for a long-term period according to the observed state.
We elaborate these key components of our proposed DRL framework, namely, \textit{state}, \textit{action}, and \textit{reward}, as follows.

\subsection{The State}
\label{sec: state}

In Tera-SpaceCom SEC networks, the DRL agent in GS perceives the following important parameters of the wireless environment as the observed state.

\begin{itemize}
   
\item
The index of each satellite (including the index of the orbital plane $n_p$ for this satellite and the index of this satellite inside this orbit $n_s$) implies its location, as well as the distance from its neighbors that affects the transmission rates.
\item
To satisfy traffic demands, the information about the amount of data in each satellite is critical.
The data amounts for each time slot are dynamic and unknown, while the mean values can be acquired through statistical analysis.
The expected value (i.e., the mean value) of data amounts to be transmitted $L_{e}$ are provided to the agent.
\item 
According to~\eqref{eq:rate}, the date rates depend on SINRs of the ISLs $\mathbf{\gamma}$, determining whether the traffic demands can be satisfied.
The channel state information can be acquired via channel estimation~\cite{chen2021hybrid}.
\item 
The agents collect which LEO satellites receive tasks before offloading and which LEO satellite is connected to the GS (i.e., the satellites offload the tasks and the satellite that is the destination of data inside the LEO network, respectively).
\end{itemize}

On account of the above factors, the observed state~\cite{RLbook} for the agent in the task offloading phase is defined as
\begin{equation}
    \label{eq: state offloading}
    \begin{aligned}
    s_{to,\tau}=&\left\{\left\{n_{p}(i),n_{s}(i),L_{e, to,\tau}(i), \mathbf{\gamma}_{to,\tau}(i), \Phi_{off,\tau}(i),\right.\right.\\
    &\left.\left.\Phi_{gs,\tau}(i)\right\}|i\in\left\{1,\dots,N\right\}\right\},
    \end{aligned}
\end{equation}
where $\Phi_{off,\tau}(i)/\Phi_{gs,\tau}(i)$ is 1 or 0, indicating whether the LEO satellite $i$ receives tasks before offloading/whether $i$ is connected to GS or not.

In contrast, in the outcome transmission phase, $\Phi_{off,\tau}(i)$ (i.e., the indicator of whether satellites offload tasks) is not considered, since the offloading process has ended. 
Hence, the observed state in the outcome transmission phase is
\begin{equation}
    \label{eq: state outcome}
    \begin{aligned}
    s_{ot,\tau}=&\left\{\left\{n_{p}(i),n_{s}(i), L_{e, ot,\tau}(i), \mathbf{\gamma}_{ot,\tau}(i),\right.\right.\\
    &\left.\left. \Phi_{gs,\tau}(i)\right\}|i\in\left\{1,\dots,N\right\}\right\}.   
    \end{aligned}
\end{equation}

\subsection{The Action}
\label{sec: action}

In the task offloading phase, we design the action $\Xi_{to, \tau}=\left\{ \mathbf{L}_{to, \tau}, \mathbf{S}_{to, \tau}, \mathbf{P}_{to, \tau} \right\}$ as follows.
\begin{itemize}
    \item In the set $\mathbf{L}_{to, \tau}=\left\{ \mathbf{L}_{to, \tau}(i),| i\in \mathbf{N}^\circ \right\}$, element $\mathbf{L}_{to, \tau}(i)$ represents the ratios of tasks assigned by the LEO satellite $i$.
    In particular,  
    $\mathbf{L}_{to, \tau}(i)=\{L_{to, \tau}(i)\} \cup \left\{ L_{to, \tau}(i,j) | j\in\mathcal{N}^i\right\}$ represents task offloading ratios for satellite $i$, as well as the neighboring satellites that connect with it.
    To ensure the task constraint in~\eqref{eq: task constraint}, 
    $L_{to, \tau}(i)+\sum_{j\in\mathcal{N}^i} L_{to,\tau}(i,j)=1$, where $i\in \mathbf{N}^\circ$, and $L_{to, \tau}(i), L_{to,\tau}(i,j)\geq0$.
    Since the allocated numbers of tasks are integers, the products are rounded up,  guaranteeing that each task can be completely processed by at least one SEC server.
    
    \item The entries of $\mathbf{S}_{to, \tau}$ denote the quotients of occupied transmitting sub-arrays as
    $\left\{ \mathbf{S}_{to, \tau}(i),| i\in \mathbf{N}^\circ \right\}$.
    More specifically, 
    $\mathbf{S}_{to, \tau}(i)=\left\{ S_{Tx, to,\tau}(i,j)| j\in\mathcal{N}^i\right\}$ represents occupied sub-array ratios for neighboring satellites of the satellite $i$.
    Due to the sub-array constraint in~\eqref{eq: subarray constraint}, 
    $\sum_{j\in\mathcal{N}^i} S_{Tx, to,\tau}(i,j)\leq 1, \forall i\in \mathbf{N}^\circ$, where $ S_{Tx, to,\tau}(i,j)\geq0$. 
    Moreover, at least one sub-array is required for the transmission.
    Hence, one sub-array is pre-allocated to each ISL.
    Then, the numbers of assigned transmitting sub-arrays are obtained by multiplying the ratios and the rest number of sub-arrays and then rounding down the products.
    
    \item In the set $\mathbf{P}_{to, \tau}$, element $ \mathbf{P}_{to, \tau}(i)=\left\{ P_{to, \tau}(i,j,k) | j\in\mathcal{N}^i,\psi_{to, \tau}(i,j,k)=1 \right\}$ is the ratios of allocated transmit power for the transmitter, i.e., the LEO satellite $i\in \mathbf{N}^\circ$.
    To guarantee the power constraint in~\eqref{eq: power constraint}, $\sum_{j\in\mathcal{N}^i,k\in\{k'|\psi_{to, \tau}(i,j,k')=1\}} P_{to, \tau}(i, j, k)\leq 1,  \forall i\in \mathbf{N}^\circ$, where $ P_{to, \tau}(i,j,k)\geq0$. 
\end{itemize}

In parallel, the action $\Xi_{ot, \tau}=\left\{ \mathbf{S}_{ot, \tau}, \mathbf{P}_{ot, \tau} \right\}$ in the outcome transmission phase is defined as follows.
\begin{itemize}
    \item $\mathbf{S}_{ot, \tau}$ is similar to $\mathbf{S}_{to, \tau}$ in the task offloading phase, which represents the ratio of occupied transmitting sub-arrays.
    However, each satellite only transmits outcome data to the next hop on the routing path.
    Hence, the element $\mathbf{S}_{ot, \tau}(i)=\left\{ S_{Tx, ot,\tau}(i,j)\right\}$, where $i\in\{1,2,\dots,N\}$, satellite $j$ is the next hop of satellite $i$, $0\leq S_{Tx, ot,\tau}(i,j)\leq1$.
    In addition, one sub-array is pre-allocated to the ISL.
    The products of ratios and the number of remaining sub-arrays are rounded down to attain the allocated numbers of sub-arrays.
    
    \item The entries in set $\mathbf{P}_{ot, \tau}$ represent the ratios of assigned transmit power for satellite $i\in\{1,2,\dots,N\}$.
    Particularly, $ \mathbf{P}_{ot, \tau}(i)=\left\{ P_{ot, \tau}(i,j,k) | \psi_{ot, \tau}(i,j,k)=1 \right\}$, where satellite $j$ is the next hop of satellite $i$, $\sum_{k\in\{k'|\psi_{ot, \tau}(i,j,k')=1\}} P_{ot, \tau}(i, j, k)\leq 1$, $P_{ot, \tau}(i,j,k)\geq0$.
\end{itemize}

\subsection{The Reward}
\label{sec: reward}

DRL is trained to provide offloading and allocation actions with the highest rewards.
For the purpose of long-term resource usage minimization, we define that instant reward for any time slot $\tau$ as a value proportional to the additive inverse of current resource occupation.
Hence, a higher reward is equivalent to lower resource usage.
In addition, the constraints are taken into account in the rewards design.
The task, power, and sub-arrays constraints in~\eqref{eq: task constraint},~\eqref{eq: power constraint}, and~\eqref{eq: subarray constraint} have been guaranteed through the definitions of actions in Sec.~\ref{sec: action}.
Therefore, the reward further includes a penalty term for latency, which is expressed as
\begin{equation}
\centering
    r_\tau=-\left(
    \chi_1 U_{\tau} + 
    \chi_2 T^{o}_\tau
    \right)
    ,
\label{eq:reward with panelty}
\end{equation}
where $\chi_1$ means a scaling factor that regulates the range of the reward for ease of the DRL convergence during DRL training.
$\chi_2$ denotes the weight of the penalty pertaining to latency.
By considering the realistic implementation of DRL training, large latency should be avoided during the whole training process, which incurs intolerable quality of service.
However, existing constrained reinforcement learning methods need a large number of training steps to mitigate the constraint violations~\cite{xu2021crpo,marchesini2022exploring,wu2020dynamic,murti2022constrained}.
Hence, instead of utilizing constrained DRL solutions, $\chi_2$ is set as a large value during the entire training process, to encourage DRL training to alleviate the large latency on the fly.
In addition, we design a piecewise penalty for latency. 
More specifically, we set a threshold for the piecewise penalty.
When the latency is below the threshold, the penalty is relatively small, to encourage DRL to increase RE without a large rise in latency.
Then, when the latency exceeds the threshold, the penalty weight is enlarged for the portion of latency that exceeds the threshold.
The large penalty can enable DRL to mitigate a large further increase in latency and, thus, preclude the latency much higher than the threshold.

Therefore, the long-term resource-efficient resource allocation problem in~\eqref{eq:total objective} is transformed into the equivalent long-term reward maximization problem in the DRL context, as
\begin{equation}
    \centering
    \label{eq:DRL problem}
    \begin{aligned}
    \mathop{\arg\max}\limits_{{\pi_p}\left(\Xi_\tau\right)}
\sum\limits_{\tau=t}^\infty\kappa^{\tau-t} \mathbb{E}_{\pi_p}\left[
r_\tau
\right].
     \end{aligned}
\end{equation}

%% file: GNNDRL.tex
\section{Graph Neural Network-Deep Reinforcement Learning Based Joint Resource Allocation and Task Offloading Algorithm}
\label{sec: GNN-DRL}

According to the framework designed in Sec.~\ref{sec: DRL description}, we propose the GRANT algorithm in this section, to solve the long-term resource allocation and task offloading problem in Tera-SpaceCom. 
To generate the continuous offloading, power, and sub-array ratios, GRANT evolves from the multi-agent deep deterministic policy gradient (MADDPG) algorithm, which is an actor-critic DRL method designed for continuous action space with multiple learning agents~\cite{lowe2017multi}.
In MADDPG algorithm, the actor provides actions by processing the observed state, while the critic evaluates the actions given by the actor for each agent via Q value (i.e., the maximal accumulated future rewards during a long-term period). 
Since only one single Q value is required to measure the reward defined in~\eqref{eq:reward with panelty}, a centralized critic is adopted in GRANT.
In addition, GNNs are deployed by the actors and critic of GRANT, to learn the relationship among LEO satellites from the connectivity structure in the LEO mega-constellation.
The GRANT framework, actor, critic, as well as algorithm are detailed as follows.

\subsection{Overall Training Framework}
\label{sec: overall training framework}

\begin{figure}[t]
\centering
        \includegraphics[width=\linewidth]{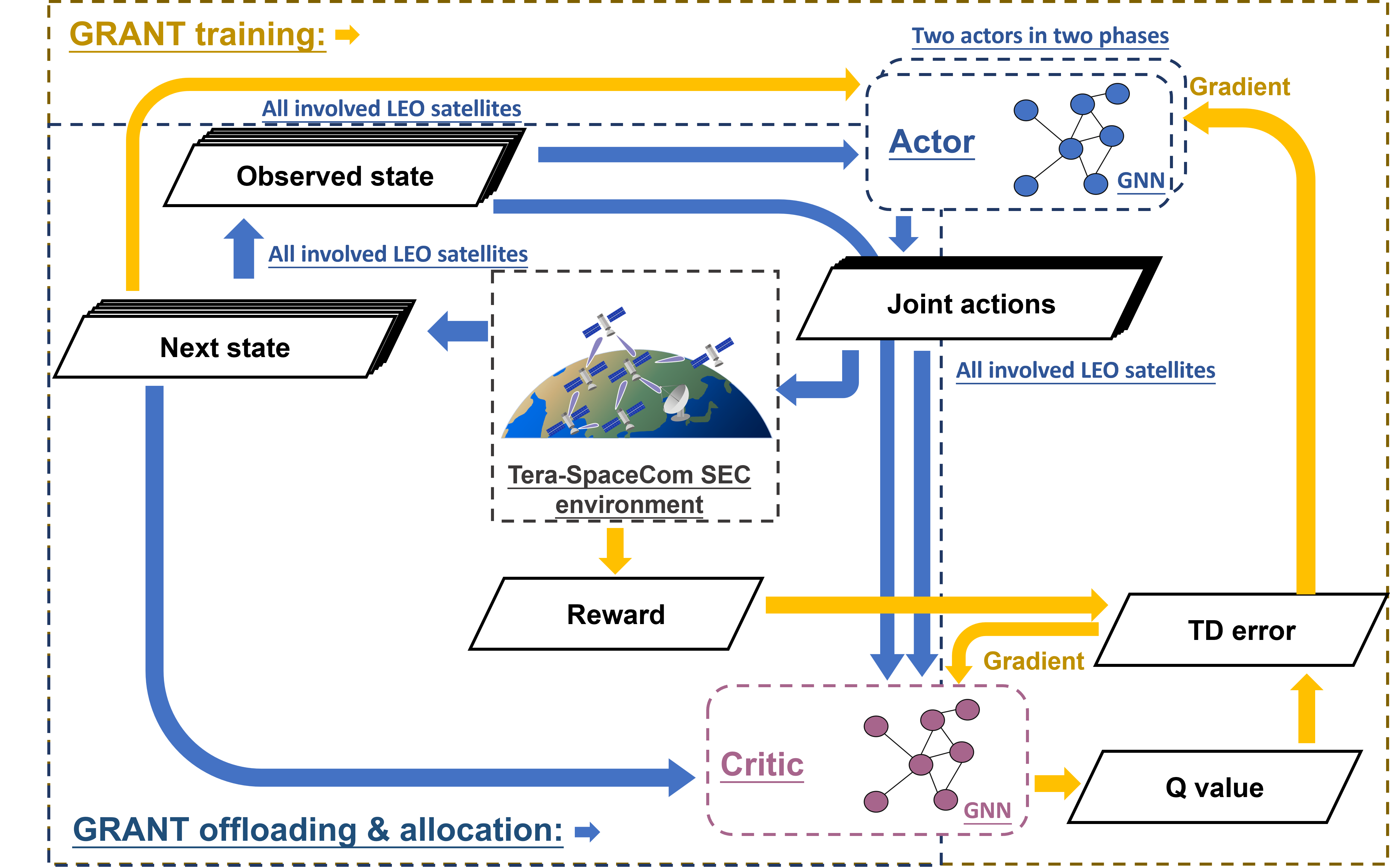} 
        \caption{{GRANT framework.}}
        \label{fig: GRANT flow}
\end{figure}

GRANT solves the proposed problem by cooperatively training two actors (designed for the agents in the task offloading phase and outcome transmission phase) and a critic, as shown in Fig.~\ref{fig: GRANT flow}. 
More specifically, by perceiving the observed states defined in Sec.~\ref{sec: state}, the actors generate the offloading and allocation actions mentioned in Sec.~\ref{sec: action}.
Then, the states and actions are processed by the critic, which outputs the Q value to assess the accumulative future rewards of offloading and allocation actions.
The experience of states, actions, and rewards are collected on the fly.
Particularly, in the task offloading phase and outcome transmission phase, the sizes of each sample of DRL experience (i.e., states, actions, and rewards) are 292 and 48 bytes under the float32 data type for each LEO satellite, respectively.  
Narrow-band feedback channel~\cite{lu2009simple} can be used to transmit this small amount of data.

In Tera-SpaceCom, the satellite that connects to the GS 
remains unchanged only in the time window 
when it is accessible to the GS (e.g., several minutes).
Based on the resource-efficient routing mentioned in Sec.~\ref{sec: satellite model}, the network topology is stable in this time window.
In each offloading and allocation training step during this time window, only one sample experience of state, action, and reward can be collected.
As a result, limited experiences can be utilized for DRL training in the time window. 
In addition, the wireless environment state is dynamic.
Therefore, on-policy training is applied in GRANT, whose training focuses on the experience of the current wireless environment.
More specifically, on-policy training works as follows:
In each training step, the current state, together with the previous state, actions, and instant reward, are input into GRANT for training.
Based on the current state and neural network parameters, the GRANT agents generate the offloading and allocation actions, which are directly leveraged in the Tera-SpaceCom SEC network.
Under the current offloading and allocation actions, the current state is updated.
Then, the next state, along with the current state, actions, and instant reward, are collected as training data for the following training step.

\subsection{Data Preprocessing}
\label{sec: data preprocessing}

\begin{figure*}[t]
\centering
        \includegraphics[width=\linewidth]{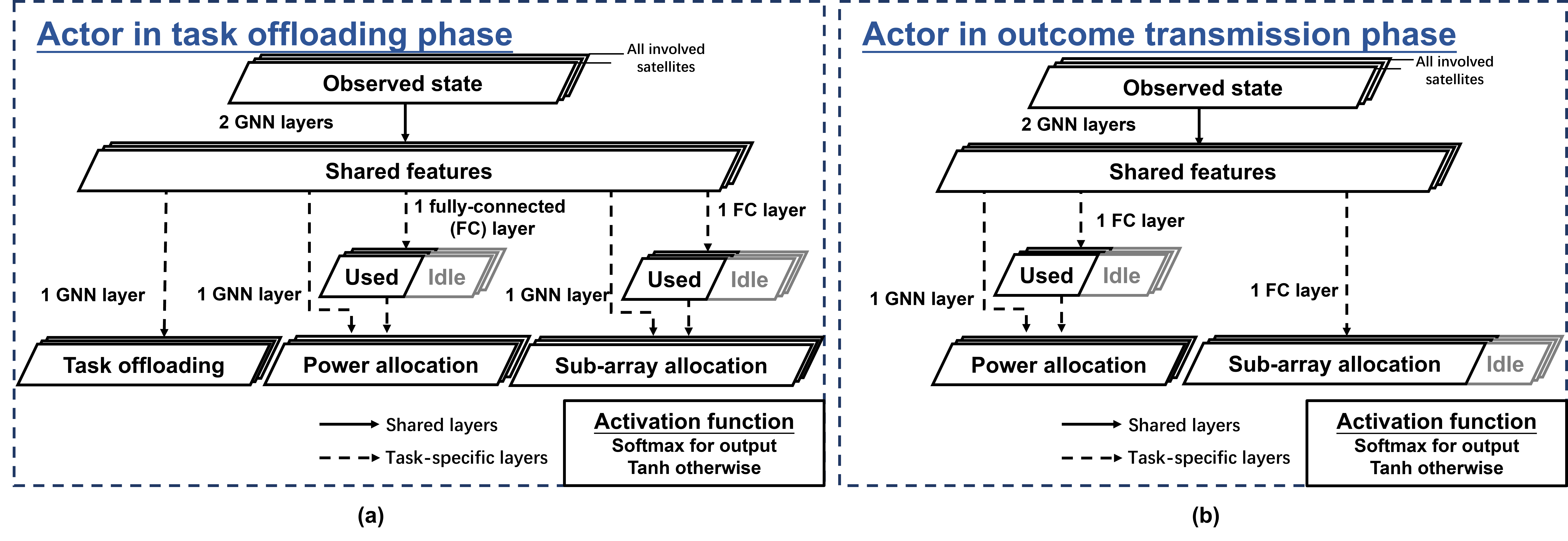} 
        \caption{{Architecture of GRANT actor (a) in the task offloading phase, and (b) in the outcome transmission phase.}}
        \label{fig: GRANT actor}
\end{figure*}

As aforementioned, the GRANT actors process the states of LEO satellites and generate offloading and allocation actions.
Instead of inputting the states from all satellites in the LEO mega-constellation, GRANT only inputs the states of the satellites involved in the computation or transmission, benefiting the DRL training from two aspects.
On the one hand, only involved LEO satellites can influence the objective in~\eqref{eq:total objective}, while the states of other LEO satellites are superfluous.
Hence, inputting the states of all LEO satellites might introduce a large amount of irrelevant information, making it challenging for DRL agents to extract useful features.
On the other hand, by pruning away the unnecessary states, the computational burden of DRL agents can be reduced. 

For the same reason, since the GRANT critic provides Q values on the basis of states and actions, the useless input states and actions are also omitted.
Particularly, the input states of the critic are the same as the input states of actors.
Since only the LEO satellites that receive tasks before offloading are required to offload unprocessed computing tasks, the actions of other satellites are discarded in the task offloading phase. 
Similarly, in the outcome transmission phase, only the actions of the involved satellites can be input into GRANT critic.

\subsection{The Actor}
\label{sec: actor}

As part of the GRANT DRL agents, GRANT actors are located at the GS as well. 
In the task offloading phase, the actor 
determines the number of tasks to be offloaded and the amount of communication resources to support task offloading, by processing the states of LEO satellites via GNN.
As illustrated in Fig.~\ref{fig: GRANT actor}(a), since the actor in the task offloading phase needs to determine the offloading ratios, as well as the transmit power and number of sub-arrays, the actor deploys the multi-task architecture.  
Specifically, the multi-task architecture is composed of shared and task-specific layers.
The shared layers extract the features from the input state defined in~\eqref{eq: state offloading} and preprocessed in Sec.~\ref{sec: data preprocessing}.
These features determine the task offloading and resource allocation of all the LEO satellites that receive tasks before offloading.
In turn, both the task offloading and resource allocation results can train the shared layers via backpropagation, encouraging the actor to identify crucial features in the observed state.
Then, based on the extracted features, the task-specific layers utilize the useful information for each task.
Consequently, the task-specific layers can output the tasks offloading as well as power and sub-array allocation at the same time, for the purpose of long-term RE maximization.
Similarly, as depicted in Fig.~\ref{fig: GRANT actor}(b), the actor in the outcome transmission phase leverages the multi-task architecture as well, which allocates the power and sub-array by inputting the state described in~\eqref{eq: state outcome} and pre-processed in Sec.~\ref{sec: data preprocessing}.

\subsection{The Critic}
\label{sec: critic}

\begin{figure*}[t]
\centering
        \includegraphics[width=\linewidth]{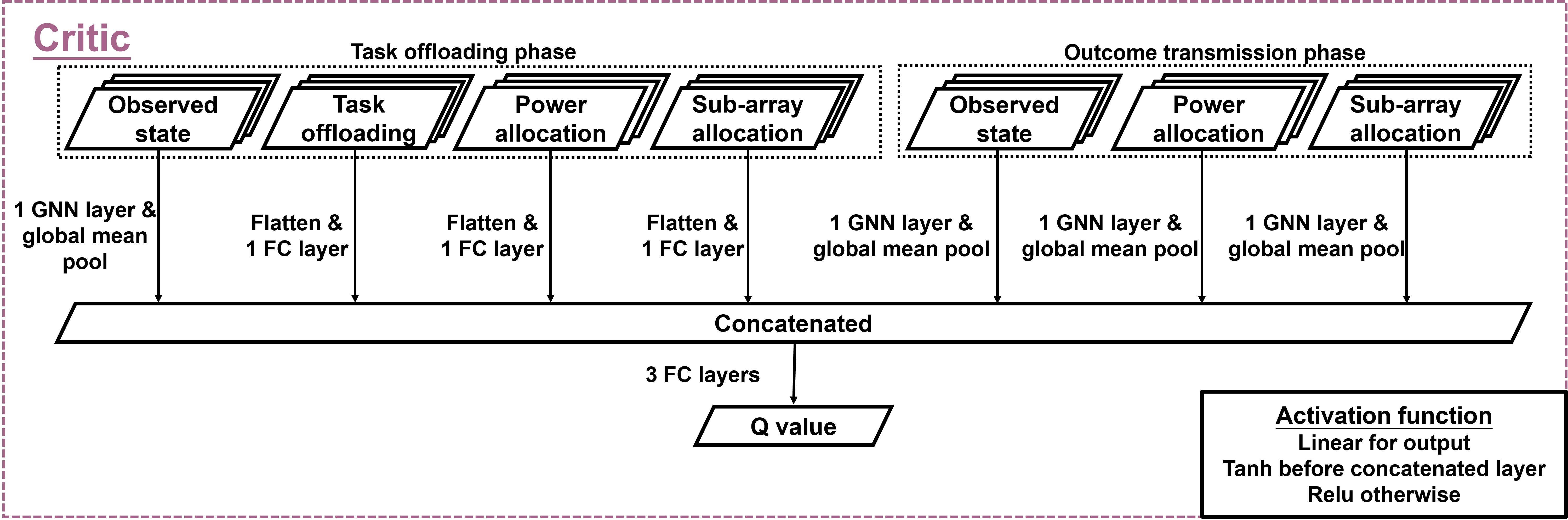} 
        \caption{{Architecture of GRANT critic.}}
        \label{fig: GRANT critic}
\end{figure*}

In addition to the actors that are trained to offload computing tasks and allocate communication resources, the critic located at GS evaluates the offloading and allocation actions provided by actors.
As shown in Fig.~\ref{fig: GRANT critic}, the critic inputs the observed states of the two actors, as well as the actions given by these actors in the pre-processed form presented in Sec.~\ref{sec: data preprocessing}.
By learning from the input states and actions, the critic provides the Q value to jointly assess the joint task offloading and resource allocation actions under the current state.
In particular, Q value represents the potential maximal future accumulated reward (i.e., additive inverse of overall resource usage) in the Tera-SpaceCom SEC network, which is given by
\begin{equation}
\label{eq:Q}
\begin{aligned}
    Q(s_{to,t},s_{ot,t},\Xi_{to,t},\Xi_{ot,t})=&\max\limits_{\left\{\pi_p(\Xi_\tau)|\tau>t\right\}}\mathbb{E}\Bigg[\sum\limits_{\tau=t}^{\infty}\kappa^{\tau-t} r_\tau|s_{to,t},\\&s_{ot,t},\Xi_{to,t},\Xi_{ot,t}\Bigg].
\end{aligned}
\end{equation}

\subsection{The GRANT Algorithm}
\label{sec: GRANT algorithm}

To learn the relationship among satellites in the LEO mega-constellation, the GRANT deploys the graph convolutional network (GCN) in its actors and critic, which is a classic GNN structure. 
Concretely, with the knowledge of the graph's connectivity, each GCN layer can aggregate and process the features from the neighbors for each LEO satellite.
The feature aggregation and processing, as well as the graph connectivity structure in the Tera-SpaceCom SEC network, are elaborated as follows.

In a graph, given the adjacent matrix $A$ and the identity matrix $I$ (with the same size as $A$), $\tilde{A}=A+I$ denotes the adjacency matrix with self-loops, which ensures that the output feature can aggregate the input features of both the node itself and its neighbors.
$\tilde{D}$ represents the diagonal degree matrix, where $\tilde{D}_{ii}=\sum_j \tilde{A}_{i,j}$.
Processed by a GCN layer, the aggregated feature matrix $F'$, whose $i^\text{th}$ row represents the features of the satellite $i$, is expressed as
\begin{equation}
\label{eq: GCN message passing}
F'= \sigma_a\left( \tilde{D}^{\frac{1}{2}} \tilde{A} \tilde{D}^{\frac{1}{2}} F \mathcal{W} \right),
\end{equation}
where $F$ means the input features, $\mathcal{W}$ symbolizes the trainable parameters, $\sigma_a(\cdot)$ is the activation function.

In the LEO mega-constellation, $A_{i,j}=1$ only when satellites $i$ and $j$ are adjacent hops in any route for task offloading or outcome transmission.
Every LEO satellite node receives data from the parent satellite node, while transmitting data to the child node.
Hence, GCN aggregates the states of each satellite node, as well as its parent and child satellite nodes.
Therefore, the LEO mega-constellation connectivity structure is an undirected graph, where $A_{i,j}=1$ if and only if $A_{j,i}=1$.

Additionally, compared with the multi-agent structure (i.e., each LEO satellite is equipped with an agent), as well as simply concatenating features of all LEO satellites, GCN utilizes the same trainable parameters $\mathcal{W}$ to process the aggregated features for each LEO satellite node.
Hence, the size of all trainable parameters of GCN remains constant regardless of the number of nodes in a graph.
GCN-based actors and critic can thereby reduce the memory burden and computational burden of DRL training, especially in the large LEO mega-constellation.

\begin{algorithm}[t]
\caption{GRANT Training.}
\KwIn{Actor network parameters in the task offloading phase: $\theta_{a,to}$; Actor network parameters in the outcome transmission phase: $\theta_{a,ot}$; Critic network parameters: $\theta_{c}$
}
\KwOut{Well-trained network parameters
}

Initialize observed state $s_{to,t},s_{ot,t}$

Initialize $\theta_{a,to},\theta_{a,ot}$ to allocate most of the resources

    \For{$\tau=t,t+1,\dots$}
    {
        Select action $\Xi_{to,\tau},\Xi_{ot,\tau}$ with random action noise according to the state $s_{to,\tau},s_{ot,\tau}$ and the policy $\pi_p$ generated by actors 
        
        Calculate reward $r_\tau$ with  $s_{to,\tau},s_{ot,\tau}$ and $\Xi_{to,\tau},\Xi_{ot,\tau}$
        
        Update $s_{to,\tau+1},s_{ot,\tau+1}$
        
        Calculate $\Xi'_{to, \tau+1}={\pi_p}(s_{to,\tau+1}|\theta_{a,to})$
        
        Calculate $\Xi'_{ot, \tau+1}={\pi_p}(s_{ot,\tau+1}|\theta_{a,ot})$
        
        Calculate $
        y_\tau=r_\tau+Q(s_{to,\tau+1},s_{ot,\tau+1},\Xi'_{to, \tau+1},\Xi'_{ot, \tau+1}|{\theta}_{c})
        $
        
        $\theta_{a,to},\theta_{a,ot}$ $\leftarrow$ Adam gradient ascent $Q(s_{to,\tau},s_{ot,\tau},\Xi_{to,\tau},\Xi_{ot,\tau}|\theta_{c})$ 
        
        $\theta_{c}$ $\leftarrow$ Adam gradient descent $
        [y_\tau-Q(s_{to,\tau},s_{ot,\tau},\Xi_{to,\tau},\Xi_{ot,\tau}|\theta_{c})]^2$
    }

\end{algorithm}

With the GNN architecture, the target of the actors is to realize the long-term RE maximization, or equivalently, to output actions with the maximal Q values.
Hence, the critic is trained to provide accurate Q values.
The accurate Q value (i.e., $Q^*$) obeys the Bellman equation as
\begin{equation}
\label{eq:bellman function}
\centering
\begin{aligned}
    &Q^*(s_{to,\tau},s_{ot,\tau},\Xi_{to,\tau},\Xi_{ot,\tau})\\=&\mathbb{E}[r_\tau+\kappa Q^*(s_{to,\tau+1},s_{ot,\tau+1},\Xi'_{to,\tau+1},\Xi'_{ot,\tau+1})|s_{to,\tau},\\&s_{ot,\tau},\Xi_{to,\tau},\Xi_{ot,\tau}],
\end{aligned}
\end{equation}
where $\Xi'_{to,\tau+1}, \Xi'_{ot,\tau+1}$ are the actions with the highest $Q^*(s_{to,\tau+1},s_{ot,\tau+1},\Xi'_{to,\tau+1},\Xi'_{ot,\tau+1})$.
As a result, the parameters $\theta_c$ of the critic should minimize the temporal difference (TD) error of Q values.
The loss function is hence defined as
\begin{equation}
        \begin{aligned}
                \label{eq:loss critic}
            Loss(\theta_c)=&\mathbb{E}\Big[\Big(r_\tau+\kappa Q(s_{to,\tau+1},s_{ot,\tau+1},\Xi_{to,\tau+1},\Xi_{ot,\tau+1}|{\theta_c}) \\&-Q(s_{to,\tau},s_{ot,\tau},\Xi_{to,\tau},\Xi_{ot,\tau}|\theta_c)\Big)^2\Big],
        \end{aligned}
        \end{equation}
where $\Xi_{to,\tau+1},\Xi_{ot,\tau+1}$ are the actions of the next training step under the current policy of the actor.
With the accurate Q value, parameters $\theta_{a,to}$ and $\theta_{a,ot}$ of the actors are trained to generate actions with the highest Q values.
Hence, the objective of the actors is to maximize the Q value, which is expressed as
\begin{equation}
    \label{eq: loss actor}
    J(\theta_{a,to})=J(\theta_{a,ot})=Q(s_{to,\tau},s_{ot,\tau},\Xi_{to,\tau},\Xi_{ot,\tau}|\theta_{c}).
\end{equation}
The training process of our proposed GRANT algorithm is summarized in Algorithm 1.

In addition, since GRANT learns on the fly with on-policy training, the bad offloading or allocation actions that cause very large latency might result in a catastrophe in Tera-SpaceCom.
Therefore, the GRANT leverages the \textit{safe initialization} and \textit{safe exploration}~\cite{hu2023deep}, which are designed for deep deterministic policy gradient (DDPG)-based resource allocation algorithms.
In particular, on the one hand, the GRANT is initialized to utilize most of the resources to ensure the actions cannot lead to a large latency in the beginning. 
On the other hand, to explore actions beyond the local optimum, Gaussian random noises are added into actions, while the sum of exploration noises for task offloading, power allocation, and sub-array allocation is forced to be zero.
After adding the noises, undesirable explored actions can be precluded, which violate the task, power, and sub-array constraints in~\eqref{eq: task constraint},~\eqref{eq: power constraint}, and~\eqref{eq: subarray constraint}.
In addition, the random noises are withdrawn if they might cause any allocated ratios to be negative, to mitigate the risk that some links are incompetent in satisfying traffic demands~\cite{hu2023deep}.

%% file: Result.tex
\section{Simulation Results and Analysis}
\label{sec: result}

In this section, experimental results are provided to evaluate the performance of the proposed GRANT algorithm pertaining to resource efficiency and latency on the fly.
These metrics are important components of the reward defined in~\eqref{eq:reward with panelty}.
Additionally, accounting for the practical implementation, metrics of interest include storage requirement and time consumption of DRL training as well.

\subsection{Simulation Parameters of Tera-SpaceCom SEC Network}
\label{sec: simulation parameters}

\begin{figure}[t] 
\centering
        \includegraphics[width=0.95\linewidth]{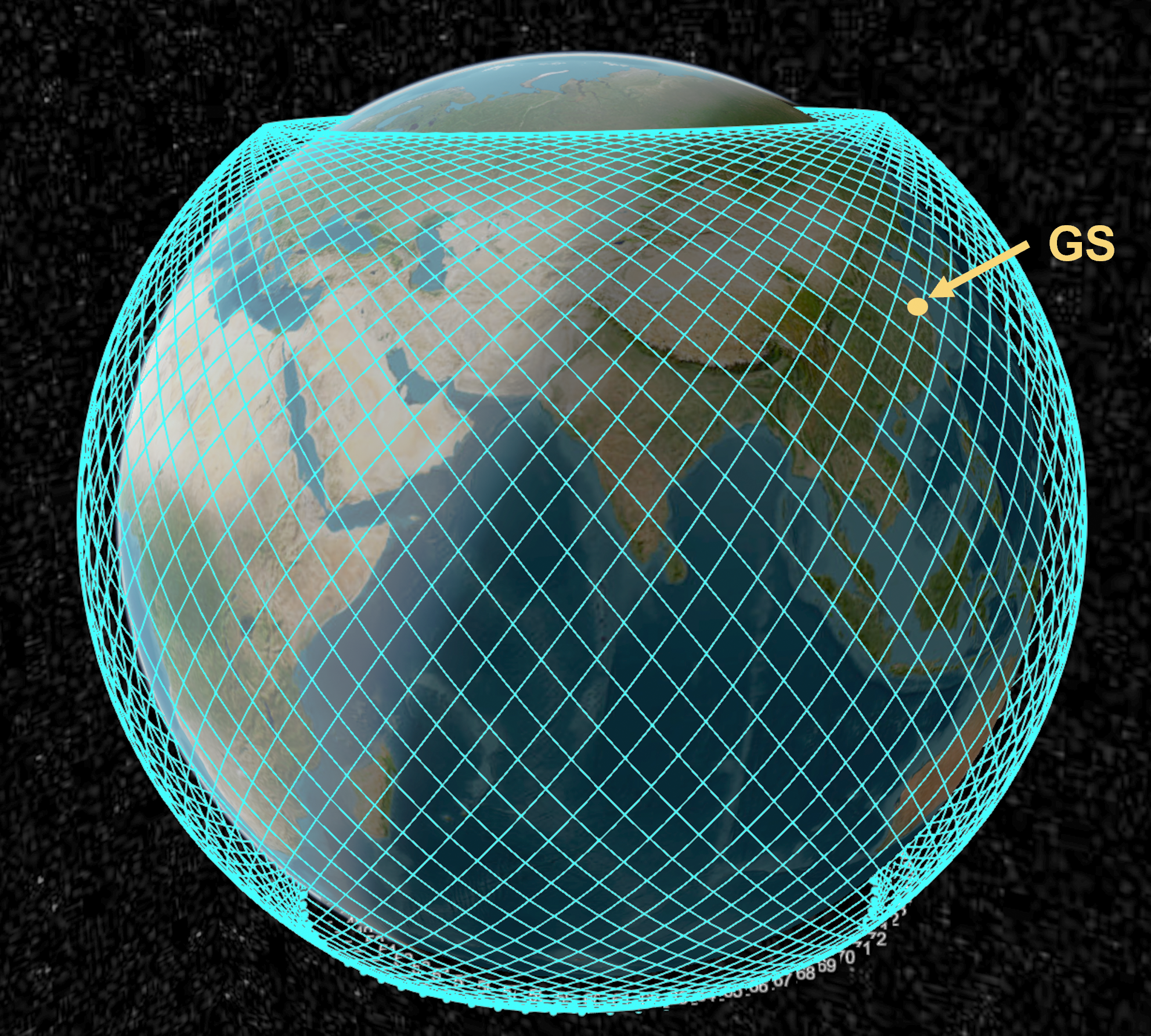} 
        \caption{LEO mega-constellation for Tera-SpaceCom SEC network.}
        \label{fig: constellation}
\end{figure}

The software environment for the simulation is established on Python 3.8.5 Pytorch 1.10.1 with AMD Ryzen ThreadRipper 3990X CPU.
As shown in Fig.~\ref{fig: constellation}, the Tera-SpaceCom SEC network in our experiments is constructed on the basis of the LEO Walker Delta mega-constellation, ensuring global services~\cite{wood2003satellite}.
Particularly, our exploited mega-constellation has the same scale as the first Starlink shell.
72 orbits comprise this mega-constellation with $53^\circ$ inclination.
Each orbit accommodates 22 satellites at an altitude of 550~km~\cite{ma2023network,kak2020online}.
Hence, this mega-constellation contains 1584 LEO satellites in total.
In addition, GS is located in Shanghai ($31.2^\circ$N, $121.4^\circ$E).
In case of large atmospheric absorption and propagation loss in the troposphere for THz satellite-ground communication, the minimal elevation angle for the GS is $15^\circ$~\cite{xing2021terahertz}.
Moreover, 10 LEO satellites that receive computing tasks before offloading are randomly selected in this mega-constellation. 
Compared to the dense LEO satellites in mega-constellation, the satellites that collect the computing tasks or relay the tasks to the LEO mega-constellation are sparser.
Hence, we assume that the selected LEO satellites are nonadjacent.

In each time slot, the mean computing demand transmitted to each LEO satellite in $\mathbf{N}^\circ$ (i.e., LEO satellites that receive computing tasks before offloading) is 2~Gbps.
The duration of each time slot is 0.05s.
We set the size of each computing task as 20~Kb~\cite{he2024age,plachy2021dynamic}.
The number of CPU cycles needed for every byte of data is $z=330$~cycles/byte~\cite{dinh2017offloading,li2020exploiting}.
Hence, the mean number of tasks for each LEO satellite in $\mathbf{N}^\circ$ is 122.
The stochastic number of tasks follows the fractional Brownian motion process, which is a commonly-used model for the data traffic~\cite{huang2021bayesian,hu2023deep}.
The SEC computing capacity for each LEO satellite $Q=2$~Gcycles/s~\cite{song2021energy,zhang2023energy}.
After computation, the ratio of the size of the outcome to the size of the unprocessed data $\beta=0.1$~\cite{chen2018joint}.  

Then, we set up parameters for THz communications of our experiments as follows.
Limited by gain of THz power amplifiers, each satellite deploys $p_\text{max}=40~\text{dBm}$~\cite{li2021propagation, you2020network,le2024resource} transmit power.
In addition, each satellite is equipped with $S_\text{max}=64$ transmitting planar sub-arrays.
Each sub-array consists of $M_x\times M_y=4\times4$ antennas with an antenna gain of $G_t=G_r=10$~dBi.
In the task offloading phase, five nonoverlapping sub-bands are selected from 130--140~GHz with 2~GHz bandwidth, while five sub-bands within 210--220~GHz are leveraged in the outcome transmission phase.

\subsection{Performance Evaluation}
\label{sec: performance}

\begin{table}[tp]
\centering
\caption{Hyperparameters in GRANT.}
\resizebox{\linewidth}{!}{
\begin{tabular}{@{}lc@{}}
\toprule
\textbf{Hyperparameter}                        & \textbf{Value}                              \\ \midrule
Attenuating factor for future rewards $\kappa$ & 0.5                                  \\
Training steps                                  & 390                                  \\
Variance of action noises                      & $5\%\times$ maximal allocated ratios \\
Learning rate                                  &  $5\times10^{-5}$ for actor, $10^{-1}$ for critic       \\
Learning rate decay ratio for actor                      & 0.95 per 3 training steps \\
Scaling factor $\chi_1$                        & 3                                  \\
Penalty weight for latency $\chi_2$     & $10/50$ for the part greater/less than 100~ms                               \\
\bottomrule
\end{tabular}
}
\label{tb:hyper}
\end{table}

The hyperparameters for the GRANT training are summarized in Table~\ref{tb:hyper}, which are tuned and finally attained to ensure the convergence of training.
Apart from the mechanisms mentioned in Sec.~\ref{sec: GNN-DRL}, we further introduce the learning rate decay that efficiently alleviates training overfitting and benefits the convergence~\cite{hu2024multi}.
As measured in our experiments, the training and action generation process of GRANT in GS contains DRL training and inference (0.044s), feeding actions back to satellites (0.089s), as well as collecting the reward of the entire network (0.144s).
Hence, the time interval for training, as well as offloading and allocation action generation, is set as 0.3s. 
As measured, the GS can communicate with its connected satellite in a time window with a duration of 117s.
Therefore, we can train the GRANT for at most 390 training steps.

On the basis of the above hyperparameters, we evaluate the performance of GRANT in terms of RE and latency on the fly, in comparison with the results of benchmark on-policy algorithms.
The benchmark algorithms are applicable for random computing demands and dynamic environments in mega-constellations with a large number of involved satellites (e.g., 315 LEO satellites, as measured in our experiment).
In particular, these benchmark algorithms can be classified into two categories, namely, GNN-based DRL methods as well as multi-agent DRL mathods.
GNN-based solutions include GNN-based actor-critic (AC) algorithm and GNN-based deep Q-learning network (DQN), while multi-agent DRL algorithms include the MADDPG, multi-agent AC (MAAC), and multi-agent DQN (MADQN).
The detailed designs of these benchmark algorithms are elaborated as follows.
\begin{itemize}
    \item GNN-based AC~\cite{sun2021graph} is a DRL structure that combines GNN with the architecture of the AC algorithm, which is designed for the stochastic selection of discrete actions.
    More specifically, the actor of AC generates the probability of choosing each action, while the critic evaluates the selected action.
    Hence, the action space is discretized.
    To avoid violating the constraint for task offloading in~\eqref{eq: task constraint}, the offloading task ratio for each neighboring satellite is picked from $\left\{0, \frac{\bar{L}}{5},\frac{2\bar{L}}{5},\dots,\bar{L}\right\}$, where $\bar{L}$ is the task ratio for one neighbor when all tasks are uniformly offloaded to the neighbors (i.e., $\frac{1}{4}$).
    The rest of the tasks are processed by the LEO satellite that receives tasks before offloading.
    Furthermore, to ensure the power and sub-array constraints in~\eqref{eq: power constraint} and~\eqref{eq: subarray constraint}, the allocated power and sub-array ratios are chosen from $\left\{0, \frac{\bar{P}}{9},\frac{2\bar{P}}{9}, \dots,\bar{P}\right\}$ and $\left\{0, \frac{\bar{S}}{9},\frac{2\bar{S}}{9}, \dots,\bar{S}\right\}$, where $\bar{P}$ and $\bar{S}$ represents the ratios of uniform power and sub-array allocation, respectively.
    In our simulations, GNN-based AC utilizes the same structure as GRANT, except for the output layer customized for it.
    \item GNN-based DQN~\cite{yang2022joint} leverages GNN in the DQN structure, which is designed to generate discrete actions in a deterministic manner. 
    Unlike DDPG-based and AC-based methods, DQN directly outputs Q values for all actions without a critic.
    Then, the action with the largest Q value is chosen in Tera-SpaceCom.
    In our simulation, GNN-based DQN exploits the same structure as the GRANT actors, except for the tailored output layer.
    The discrete action space is the same as the aforementioned GNN-based AC.
    \item MADDPG~\cite{yang2023cooperative} is a deterministic DRL algorithm that generates continuous actions in multi-agent environments.
    Different from GRANT, MADDPG deploys a unique actor for each LEO satellite involved in the task offloading phase and outcome transmission phase without the assistance of GNN.
    In our experiments, each agent of MADDPG uses the same architecture and action space as GRANT, while fully-connected (FC) layers are utilized to replace the GNN in GRANT.
    \item Similar to MADDPG, MAAC~\cite{araf2022uav} employs an AC agent for each LEO satellite without GNNs.
    Furthermore, MAAC adopts an attention mechanism in critic that figures out the importance of actions of all agents, for ease of cooperative training~\cite{iqbal2019actor}.
    Since one Q value is needed for the reward in~\eqref{eq:reward with panelty}, we use a centralized critic for all agents.
    MAAC exploits FC layers to replace GNN in GNN-based AC, while leveraging the same action space as GNN-based AC.
    \item MADQN~\cite{waqar2022computation} deploys a DQN agent for each LEO satellite involved in the computation and transmission processes.
    Since actor-critic architecture is not adopted in the DQN-based structure, the MADQN agents are separately trained without a centralized critic. 
    In addition, MADQN has the same architecture and action space as GNN-based DQN, except for the GNN layers that are replaced by the FC layers.
\end{itemize}

\begin{figure}[t] 
\centering
        \includegraphics[width=\linewidth]{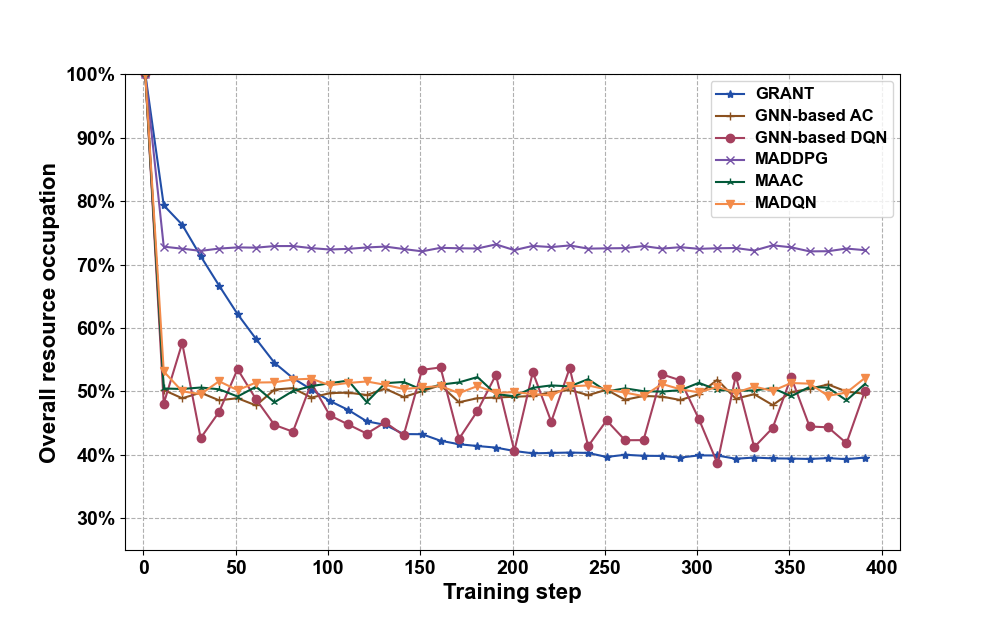} 
        \caption{Resource usage comparison.}
        \label{fig: resource}
\end{figure}
\begin{figure}[t] 
\centering
        \includegraphics[width=\linewidth]{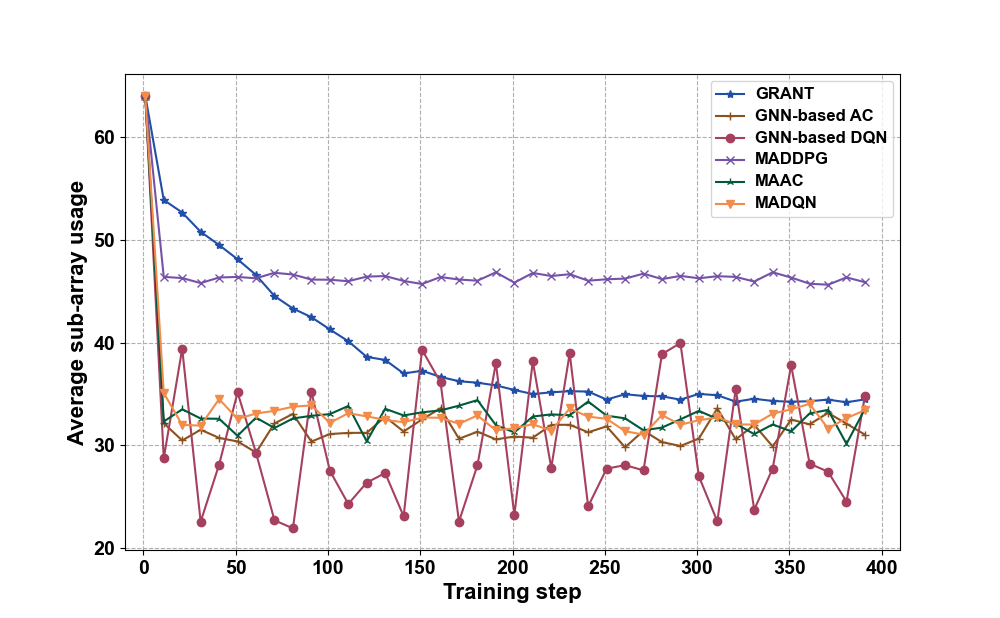} 
        \caption{Sub-array usage comparison.}
        \label{fig: array}
\end{figure}
\begin{figure}[t] 
\centering
        \includegraphics[width=\linewidth]{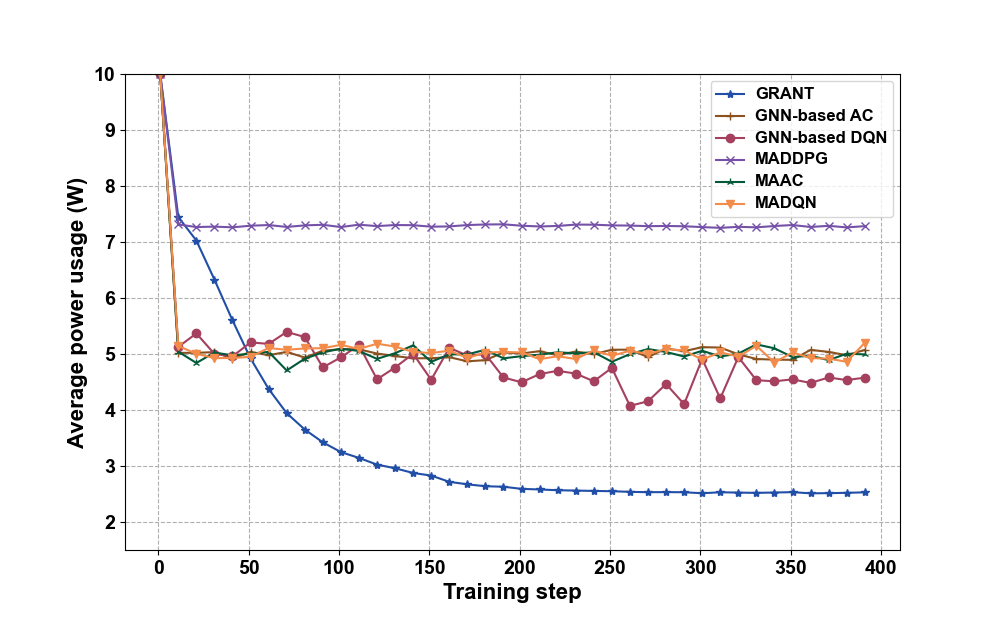} 
        \caption{Power usage comparison.}
        \label{fig: power}
\end{figure}

Fig.~\ref{fig: resource} illustrates the resource usage during the on-policy training process on the fly. 
GRANT substantially reduces the ratio of resource utilization during training, which is equivalent to improving RE.
GRANT converges at 40\% resource occupation ratio, which is substantially lower than 50\% resource usage for GNN-based AC, MAAC, and MADQN, as well as 73\% resource usage for MADDPG. 
MADDPG has the highest resource usage, since it can utilize most of the resources to avoid large latency initially via the initialization mechanism designed for DDPG structures, which is also used in GRANT.
Even if GNN-based DQN can realize 40\% resource usage as well, it cannot converge within the limited training time and frequently reaches resource usage 10\% higher than GRANT.
More specifically, the sub-array and power usage are shown in Fig.~\ref{fig: array} and Fig.~\ref{fig: power}, whose trends are similar to the overall resource usage.
GRANT occupies 34 sub-arrays (averaged over all involved satellites) after convergence, which is comparable to the GNN-based AC, MAAC, MADQN, and remarkably less than MADDPG (i.e., 46 sub-arrays).
Unconverged GNN-based DQN fluctuates in the range of $[22, 40]$.
Moreover, the power of GRANT converges at 2.5~W averaged over all involved satellites, which is much less than 7.3~W of MADDPG and 5~W of remaining algorithms.
Unlike GRANT, the solutions for discrete actions and the multi-agent methods cannot efficiently reduce sub-array and power usage in the complex Tera-SpaceCom SEC network with massive LEO satellites.

\begin{figure}[t] 
\centering
        \includegraphics[width=\linewidth]{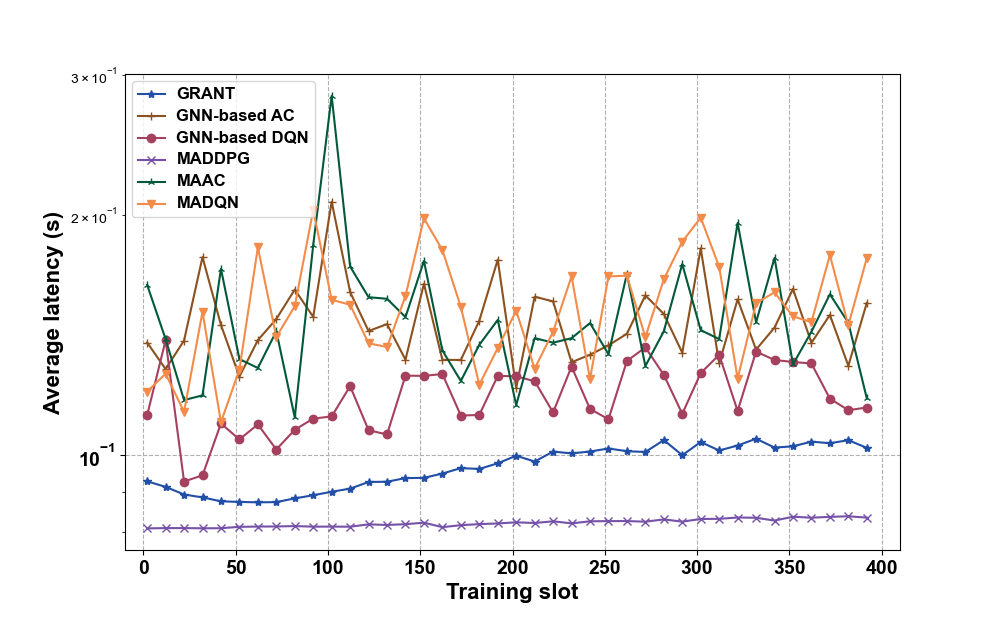} 
        \caption{Average latency comparison.}
        \label{fig: latency}
\end{figure}

\begin{figure}[t] 
\centering
        \includegraphics[width=\linewidth]{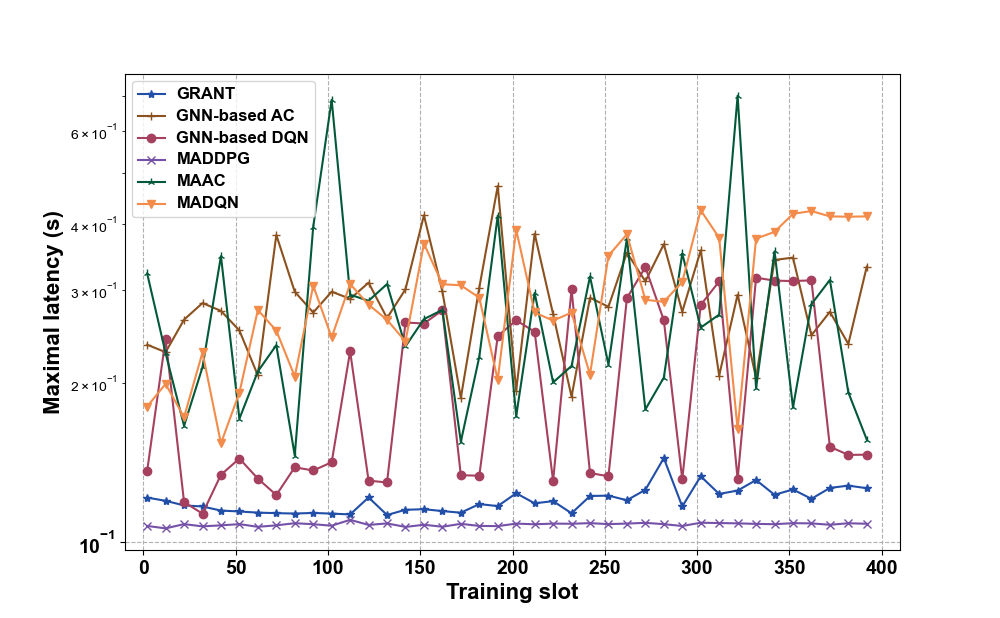} 
        \caption{Maximal latency comparison.}
        \label{fig: max latency}
\end{figure}

\begin{figure}[t] 
\centering
        \includegraphics[width=\linewidth]{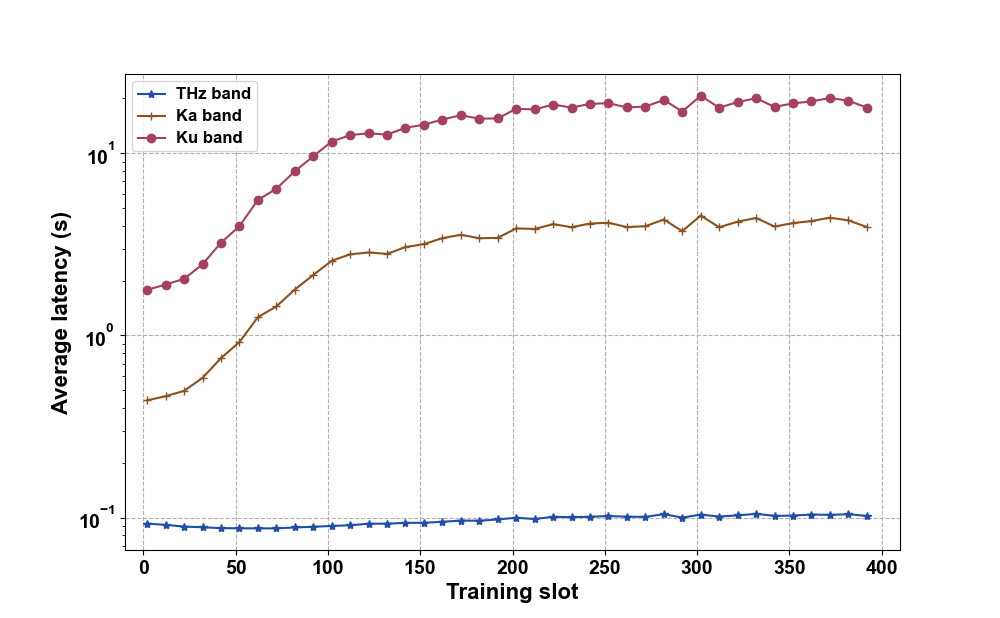} 
        \caption{Average latency for different frequency bands.}
        \label{fig: low freq latency}
\end{figure}

\begin{figure}[t] 
\centering
        \includegraphics[width=\linewidth]{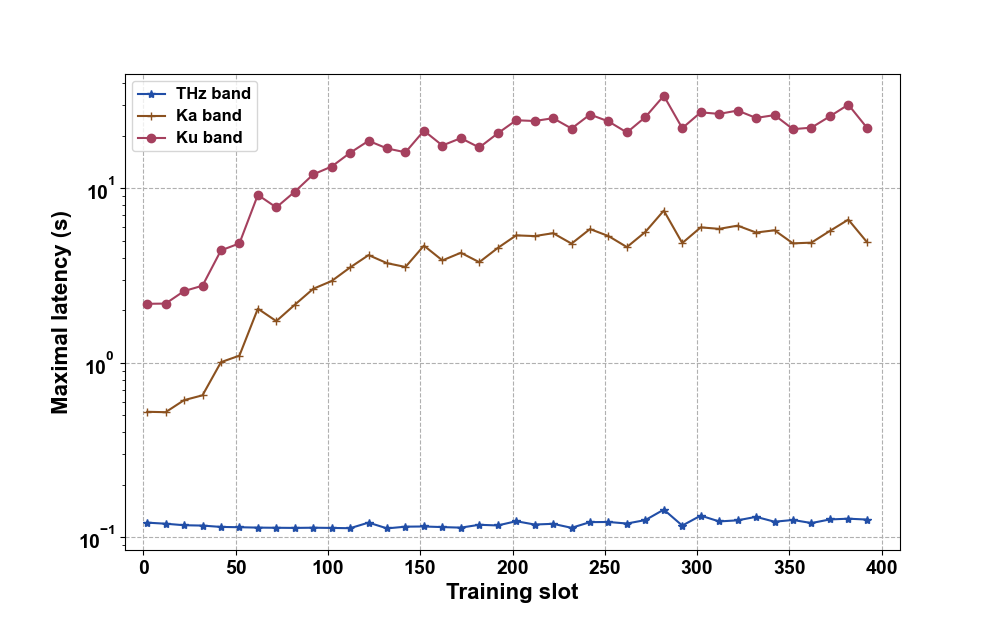} 
        \caption{Maximal latency for different frequency bands.}
        \label{fig: low freq max latency}
\end{figure}

Apart from the RE, to guarantee the quality of service, low latency is significant as well.
Fig.~\ref{fig: latency} demonstrates latency of the computing tasks averaged over LEO satellites in $\mathbf{N}^\circ$.
GRANT not only enhances the RE, but reduces the latency from 93~ms to 87~ms in the beginning, benefitting from the penalty of latency in the reward.
Then, with the further increase in RE, the latency of GRANT increases.
Furthermore, when the latency exceeds the threshold of 100~ms after about $200$ training steps, the resource usage of GRANT converges with a latency of 105~ms. 
In comparison, MADDPG achieves the lowest latency (i.e., 84~ms) at the expense of 83\% more resource usage compared to GRANT.
Other algorithms result in up to 35--178~ms more latency in the Tera-SpaceCom SEC network with lower RE than GRANT.
Moreover, as illustrated in Fig.~\ref{fig: max latency}, we also measure the maximal latency in the Tera-SpaceCom SEC network, which relates to the quality of services for the farthest LEO satellite in $\mathbf{N}^\circ$.
In addition, the maximal latency equals the delay when DRL algorithm receives the reward and, thus, influences the time interval between consecutive training steps.
A longer time interval incurs a longer time for action explorations, experience collections, and convergence for DRL algorithms.
With the exception of MADDPG (i.e., 110~ms by sacrificing RE), GRANT realizes a maximal latency of 144~ms, outperforming other benchmark algorithms (i.e., more than 330~ms).

We further compare the latency of THz communications and the classical low-frequency Ku band and Ka band communications.
More specifically, the central frequencies for the task offloading phase and the outcome transmission phase are 14~GHz/30~GHz and 16~GHz/35~GHz for the Ku/Ka band, respectively.
THz, Ku, and Ka bands all apply the same resource usage and task offloading given by the GRANT algorithms.
In addition, the Ku and Ka bands utilize the same fractional bandwidth and effective aperture of antenna array as the THz band~\cite{nie2021channel}.
As illustrated in Fig.~\ref{fig: low freq latency} and Fig.~\ref{fig: low freq max latency}, unlike the THz band, the latency of the Ka/Ku band dramatically increases with the decrease in resource usage. 
The Ka/Ku band results in the 4.55s/20.68s average latency, as well as 7.46s/33.95s maximal latency.
Remarkably, the THz band realizes $\frac{1}{43}$ and $\frac{1}{197}$ of the average latency, as well as  $\frac{1}{52}$ and $\frac{1}{235}$ of the maximal latency, compared to the Ka and Ku bands, respectively.

\begin{figure}[t] 
\centering
        \includegraphics[width=\linewidth]{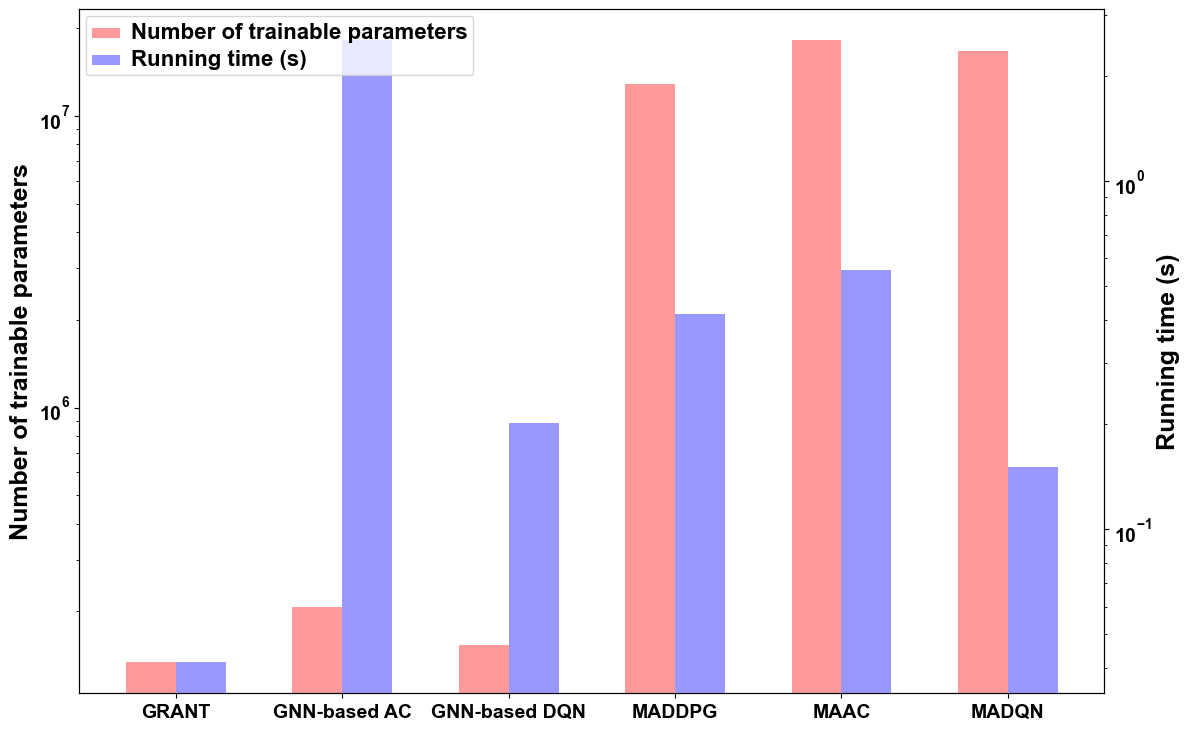} 
        \caption{Number of trainable parameters and running time comparison.}
        \label{fig: parameter number and time}
\end{figure}

Last but not least, by considering the practical deployment of DRL algorithms, we evaluate the number of trainable parameters and running time, as shown in Fig.~\ref{fig: parameter number and time}.
The number of trainable parameters affects the memory burden of the DRL algorithm.
GRANT uses the $1.3\times10^5$ trainable parameters, which is less than $2.1\times10^5$ of GNN-based AC and $1.5\times10^5$ GNN-based DQN.
The reason is that GRANT only exploits one output neuron to directly generate each output value, while GNN-based AC and DQN select each output value from multiple discrete values given by multiple output neurons.
In addition, each LEO satellite employs a separate actor for multi-agent methods. 
Hence, compared to the GNN-based methods, MADDPG, MAAC, and MADQN have much larger numbers of trainable parameters to process the states and actions of all involved satellites with FC layers (i.e., $1.3\times10^7$, $1.8\times10^7$, and $1.7\times10^7$, respectively).
The amount of trainable parameters in GNN-based algorithms is just about one percent of that in multi-agent solutions, revealing the advantage of GNN-based DRL in terms of memory efficiency.

In addition, the running time relates to the training speed, as well as the computation burden of DRL agents.  
Without the process of discrete offloading and allocation ratio selection from output neurons, GRANT spends 0.044s for each training step, which is much faster than AC-based and DQN-based solutions (i.e., 2.53s of GNN-based AC, 0.202s of GNN-based DQN, 0.556s of MAAC, and 0.151s of MADQN).
Since actors of multi-agent methods can expedite running speed by learning and selecting discrete actions in a parallel manner, the running time of MAAC and MADQN is shorter than GNN-based AC and GNN-based DQN, respectively.
However, different from GRANT, the critic of MADDPG needs a large number of extra parameters in FC layers without the assistance of GNN to process all the input states and actions, which is time-consuming.
In addition, DDPG architecture directly outputs actions, without the discrete value selection process that can be accelerated by multi-agent architecture. 
As a result, the running time of MADDPG per training step is 0.414s, which is substantially larger than that of GRANT as well.

%% file: conclusion.tex
\section{Conclusion}
\label{conclusion}
In this paper, we proposed the GRANT algorithm to maximize the long-term RE via task offloading and resource allocation in the SEC network of Tera-SpaceCom. 
More specifically, GNNs are deployed in the DRL architecture, to learn the relationships among LEO satellites based on their connectivity structure in the mega-constellation.
Furthermore, the exploitation of multi-agent and multi-task architecture empowers the cooperative training of computing task offloading and communication resource allocation of all involved LEO satellites.

Our simulation results illustrate that GRANT achieves the lowest communication resource occupation, or equivalently, the highest RE with a latency of 105~ms.
In contrast, the benchmark solutions either have more than 33\% higher latency or achieve 20\% less latency at the cost of 83\% more resource usage.
Additionally, THz communications achieve only $\frac{1}{43}$ and $\frac{1}{197}$ of the latency, as well as $\frac{1}{52}$ and $\frac{1}{235}$ of the maximum latency of low-frequency Ka and Ku band communications, respectively.
Moreover, the GRANT algorithm outperforms benchmark algorithms regarding the lower number of trainable parameters and faster running time, revealing its superior memory efficiency and computational efficiency. 